\documentclass{article}

\usepackage[preprint]{neurips_2025}

\usepackage[utf8]{inputenc} 
\usepackage[T1]{fontenc}    
\usepackage[colorlinks=true]{hyperref}       
\usepackage{url}            
\usepackage{booktabs}       
\usepackage{amsfonts}       
\usepackage{nicefrac}       
\usepackage{microtype}      
\usepackage{graphicx}       
\usepackage{amsmath}        
\usepackage{multirow}
\usepackage[table,xcdraw]{xcolor}
\usepackage{caption}
\usepackage{float}
\usepackage{makecell}

\definecolor{reason}{HTML}{4F81BD}
\definecolor{answer}{HTML}{9BBB59}
\definecolor{search}{HTML}{C0504D}
\definecolor{information}{HTML}{F79646}
\definecolor{lightgreen}{HTML}{F5FFFA}
\newcommand{\reason}[1]{\textcolor{reason}{#1}}
\newcommand{\answer}[1]{\textcolor{answer}{#1}}
\newcommand{\search}[1]{\textcolor{search}{#1}}
\newcommand{\information}[1]{\textcolor{information}{#1}}

\title{MMSearch-R1: Incentivizing LMMs to Search}

\author{
  Jinming Wu$^{1,*}$,
  Zihao Deng$^{1,*}$,
  Wei Li$^{1}$, 
  Yiding Liu$^{1}$, 
  Bo You$^{1}$,\\
  \textbf{Bo Li}$^{2,\dagger}$,
  \textbf{Zejun Ma}$^{1}$,
  \textbf{Ziwei Liu}$^{2}$ \\
  $^{1}$ByteDance \;
  $^{2}$S-Lab, NTU \\
  \texttt{\textcolor{blue}{
  \href{https://github.com/EvolvingLMMs-Lab/multimodal-search-r1}{https://github.com/EvolvingLMMs-Lab/multimodal-search-r1}}}
}

\begin{document}

\renewcommand{\thefootnote}{\fnsymbol{footnote}}
\footnotetext{$*$ Equal Contribution;~$\dagger$ Work collaborated with ByteDance}
\renewcommand{\thefootnote}{\arabic{footnote}}

\maketitle


\vspace{-3mm}
\begin{figure}[h]
  \centering
  \vspace{-10pt}
  \includegraphics[width=0.96\textwidth]{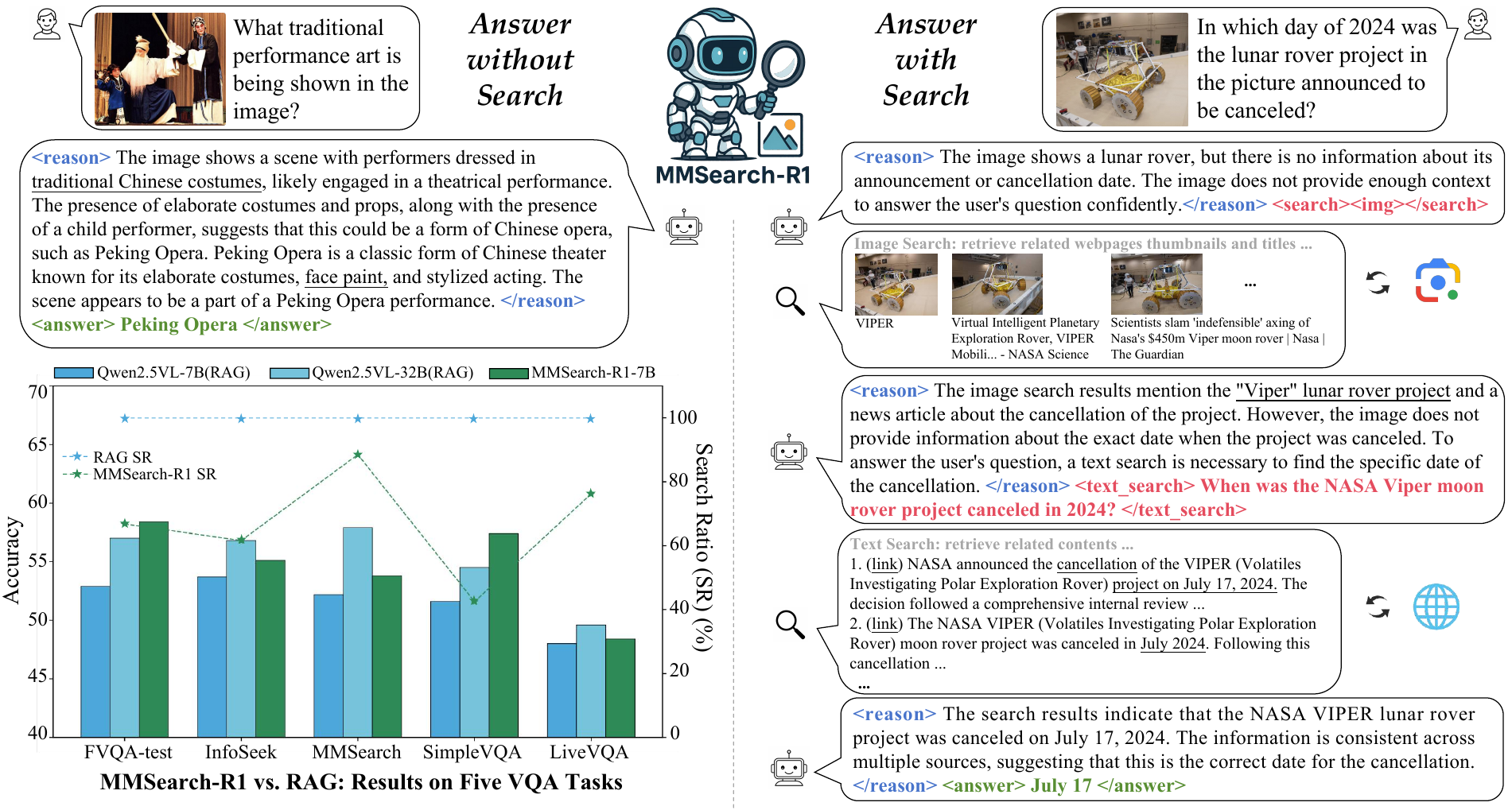}
  \caption{\textbf{Overview of MMSearch-R1.} MMSearch-R1 learns to recognize the boundaries of its knowledge and perform on-demand search, significantly reducing the number of searches required while outperforming RAG-based models on knowledge-intensive and info-seeking VQA tasks.
}
  \label{fig:page1}
\end{figure}

\begin{abstract}
  Robust deployment of large multimodal models (LMMs) in real-world scenarios requires access to external knowledge sources, given the complexity and dynamic nature of real-world information. Existing approaches such as retrieval-augmented generation (RAG) and prompt engineered search agents rely on rigid pipelines, often leading to inefficient or excessive search behaviors. We present \textbf{MMSearch-R1}, the first end-to-end reinforcement learning framework that enables LMMs to perform on-demand, multi-turn search in real-world Internet environments. Our framework integrates both image and text search tools, allowing the model to reason about when and how to invoke them guided by an outcome-based reward with a search penalty.
  To support training, We collect a multimodal search VQA dataset through a semi-automated pipeline that covers diverse visual and textual knowledge needs and curate a search-balanced subset with both search-required and search-free samples, which proves essential for shaping efficient and on-demand search behavior.
  Extensive experiments on knowledge-intensive and info-seeking VQA tasks show that our model not only outperforms RAG-based baselines of the same model size, but also matches the performance of a larger RAG-based model while reducing search calls by over \textbf{30\%}. 
  We further analyze key empirical findings to offer actionable insights for advancing research in multimodal search.
\end{abstract}

\newpage
\section{Introduction}
\label{main/intro}
Scaling up visual-text pair data by leveraging large-scale, high-quality, and diverse datasets across different training stages \cite{dataset/clip,dataset/mint-1t,dataset/2p5-years-in-class,dataset/100b-pretrain,dataset/sail-douyin} has become a fundamental paradigm for acquiring grounded knowledge of the visual world, driving breakthroughs in Large Multimodal Models (LMMs) \cite{lmm/gpt4o,lmm/gemini,lmm/kimi-vl,lmm/qw25vl,lmm/llava-ov,lmm/aira,lmm/vila,lmm/internvl}. This foundation enables strong performance across a wide range of visual understanding tasks, such as visual question answering and captioning, facilitating the transition of LMMs from research prototypes to practical, intelligent assistants in everyday life.
 
 However, this paradigm faces inherent limitations when handling complex and dynamic real-world knowledge \cite{intro/pretrain-limit-1,intro/pretrain-limit-2}. Specifically, long-tail information such as facts that emerge after the model’s training cut-off or domain-specific knowledge constrained by privacy, copyright, or security considerations is difficult to capture through static training alone. As a result, leading LMMs continue to struggle with knowledge-intensive and information-seeking VQA tasks where external and up-to-date knowledge is often required \cite{intro/struggle-1,vision-rag/mmsearch,dataset/livevqa}. When confronted with inputs beyond their internal knowledge boundaries, such as unfamiliar visual content or previously unseen textual information, models are prone to hallucinations \cite{intro/hallucination-risk-2,intro/hallucination-risk-3}, which severely compromise their reliability in applications that demand factual accuracy and trustworthiness \cite{intro/hallucination-risk-1}.

Integrating the ability to interact with search tools into LMMs to search for external information offers a promising solution to these limitations \cite{vision-rag/mmsearch,vision-rag/avis,vision-rag/vsa}. Existing approaches can be broadly categorized into two paradigms: (1) Retrieval-Augmented Generation (RAG) \cite{vision-rag/murag,vision-rag/visrag,vision-rag/rag-vl}, which retrieves external visual or textual information to guide model generation; and (2) Prompt-engineered agents \cite{vision-rag/avis,vision-rag/vsa,vision-rag/mmsearch}, which prompt large models to iteratively perform web searches and reason over the retrieved results to generate answers. However, these approaches remain suboptimal in practice. RAG-based methods follow a fixed retrieve-then-generate workflow grounded in static knowledge bases, often resulting in over-retrieval, high computational cost, and the unrealistic assumption that all required information is already present in the corpus. Such a controlled setting fails to capture the dynamic and unpredictable nature of real-world scenarios, rendering these systems vulnerable in practical deployments. On the other hand, prompt-engineered agents interact with real-world search engines, but the model parameters are not optimized through learning. As a result, the models do not truly learn how to interact effectively with search tools or adapt their behavior to open world environments. This motivates the development of methods that teach models to search on demand and interact effectively with search tools, ensuring practical usability in dynamic real-world settings.

Recent advances, such as OpenAI's o series \cite{rl/openai-o1,openai-o3_o4_mini_syscard} and DeepSeek-R1 \cite{rl/deepseek-r1}, have highlighted the potential of end-to-end reinforcement learning (RL) to enhance the reasoning capabilities of large-scale models. In addition, OpenAI introduced the Deep Research \cite{openai-deep-research}, claiming that training models via end-to-end RL to interact with search tools and web content can significantly improve their ability to solve complex open-ended tasks that require iterative reasoning and information seeking. In the open-source community, efforts such as DeepResearcher \cite{text-search/deepresearcher}, Search-R1 \cite{text-search/search-r1}, and ReSearch \cite{text-search/ReSearch} have followed this direction, applying end-to-end RL to improve models’ abilities in multiturn search and retrieval-augmented generation, aiming to boost performance on information-seeking question answering tasks. However, existing work mainly focuses on text-based search, while in the multimodal domain, challenges such as constructing suitable training data and designing RL frameworks to incentivize models to perform search in real-world environments remain underexplored.

In this work, we focus on training LMMs to learn three key search-related abilities: (1) \textit{when to search}, (2) \textit{what to search for}, and (3) \textit{how to reason over search results to answer user queries}. By exploring these questions, we propose \textbf{MMSearch-R1}, the first end-to-end RL-based solution designed to equip LMMs with the capability to perform search on demand in real-world internet environments. Our efforts can be summarized as follows:

\begin{itemize}
    \item \textbf{Datasets Construction} We propose an automated method for constructing a multimodal search VQA dataset by estimating the model's familiarity with the visual and textual knowledge required to answer each question. Based on this estimation, we generate a mixture of search-required and search-free VQA samples, which is crucial for shaping the model’s ability to perform on-demand search. In addition, we complement the dataset with manually annotated test data that covers diverse knowledge categories and a range of difficulty levels, ensuring a comprehensive evaluation of the model's performance.

    \item \textbf{Multimodal Search Tool Integration} We build a real-world search pipeline consisting of two tools: an image search tool, which allows the model to retrieve webpage thumbnails and titles related to a user-provided image to help identify unfamiliar visual content; and a text search tool, which enables the model to issue precise textual queries to retrieve relevant webpages and acquire textual knowledge.

    \item \textbf{Better Performance through Wiser Search} We demonstrate that by incorporating an outcome-based reward with search penalty, the GRPO algorithm \cite{rl/deepseekmath} can be directly applied to LMMs without cold-start initialization. This setup encourages models to perceive their knowledge boundaries and then perform search when necessary. As a result, the model learns to reason about when and how to execute multi-turn, complex search strategies. In knowledge-intensive and information-seeking VQA tasks, MMSearch-R1-7B not only outperforms RAG-based baselines of the same model size, but also achieves competitive performance compared to a 32B RAG-based model, while significantly reducing the number of search calls by over \textbf{30\%}.

    \item \textbf{Open-Sourcing Data and Training Framework} We conduct extensive experiments to derive key empirical findings, which we share with the community to provide deeper insights into search-augmented multimodal reasoning. In addition, we will open source our data and the complete training framework to facilitate further research in this area.
\end{itemize}
\section{Building Iterative Multimodal Search-Integrated RL Framework}
\subsection{Group Relative Policy Optimization (GRPO)}
As shown in the top part of Figure~\ref{fig:grpo_search_rollout}, we adopt standard GRPO as our base RL algorithm, with modifications to allow search interactions with the real-world environment during the rollout process. Originally introduced in DeepSeekMath \cite{rl/deepseekmath}, GRPO is a variant of the Proximal Policy Optimization (PPO) algorithm \cite{rl/ppo}. Unlike PPO, GRPO estimates the baseline directly from a group of rewards, without relying on a value function, which significantly reduces the computational burden during training. The details of the GRPO algorithm are provided in the Appendix~\ref{appendix:grpo}.

\subsection{Multimodal Search Tools} 
\begin{figure}[t]
  \centering
  \includegraphics[width=0.96\textwidth]{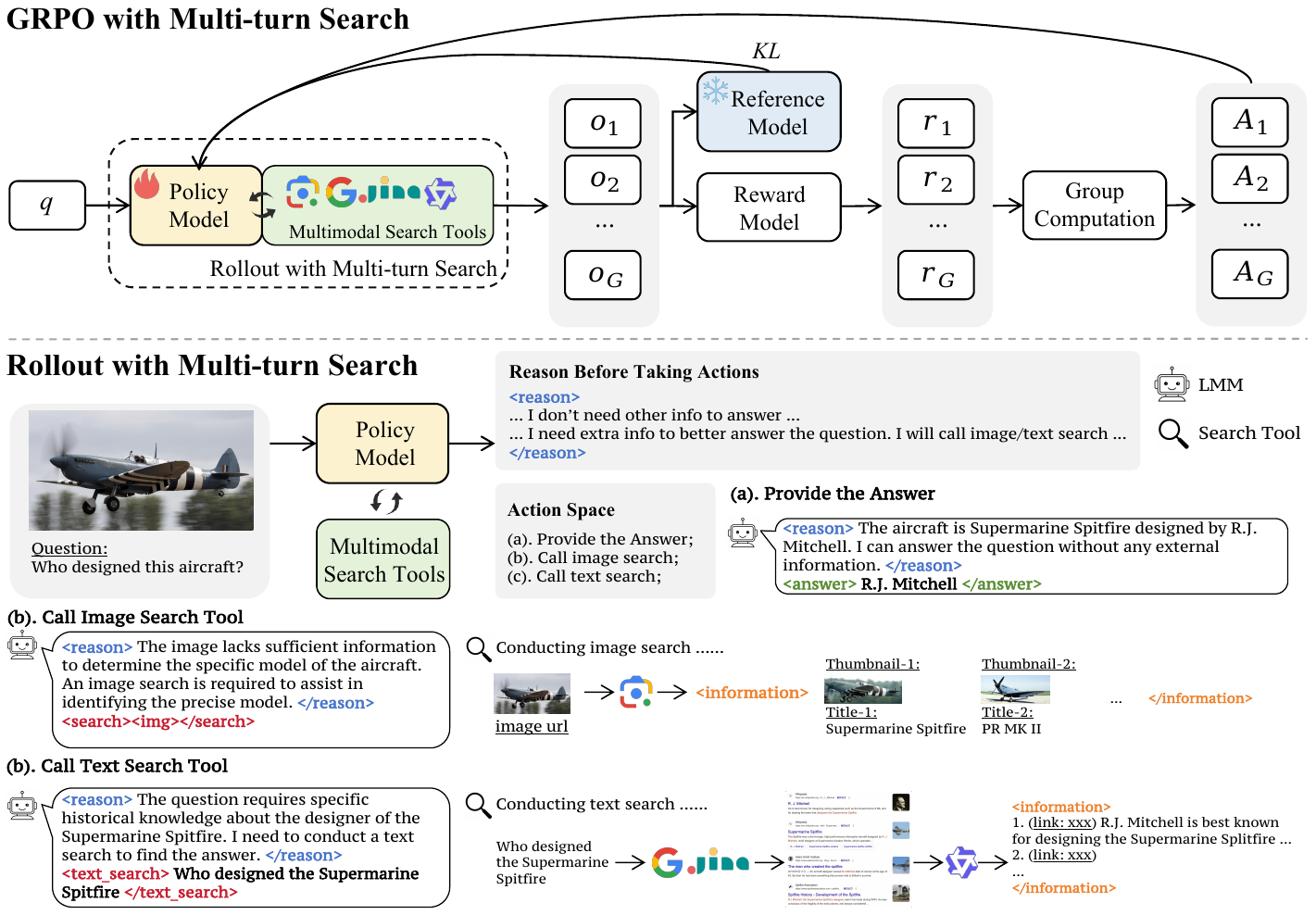}
  \caption{Illustration of training in MMSearch-R1. Top: The GRPO training pipeline integrated with multimodal search tools. Bottom: A detailed view of the rollout process and search tool execution.}
  \label{fig:grpo_search_rollout}
\end{figure}

A fully functional search toolkit is crucial for solving information-seeking VQA tasks. As illustrated in the bottom part of Figure~\ref{fig:grpo_search_rollout}, we equip the model with two types of search tools for interacting with real-world internet content. The first is an image search tool powered by SerpApi.\footnote{SerpApi: https://serpapi.com/} The model submits the image from the original question to the image search engine, which returns the top-5 visually matched webpages in an interleaved format, each represented by a thumbnail and a title. This helps the model identify the key visual entities present in the input image. The second is a text search pipeline composed of SerpApi, Jina Reader\footnote{Jina Reader: https://jina.ai/reader/}, and a webpage summarizer. The model autonomously generates a text query related to the original question and submits it to the search tool to retrieve relevant information. SerpApi returns the top-5 webpage URLs associated with the query. Jina Reader then fetches the full content of each webpage and converts it into a clean, readable format. A summarization model (Qwen3-32B \cite{qwen3-tech-report} in our implementation) processes each page and extracts the content most relevant to the user’s question into a concise summary. Implementation details can be found in the Appendix~\ref{appendix:mm-search-tools}.

\subsection{Rollout with Multi-turn Multimodal Search}
To address complex information-seeking tasks, the model may engage in multiple rounds of interaction with the internet environment before providing a final answer. As illustrated in the bottom part of Figure~\ref{fig:grpo_search_rollout}, the rollout process is multi-turn and iterative. We structure the prompt to guide the model to perform reasoning within the \reason{<reason>} and \reason{</reason>} tags whenever new information is received, including the original question or any retrieved search results, and then choose an action from a predefined action space (see Appendix~\ref{appendix:train-prompts} for the full prompt). If the model decides to answer the question, it is required to place its response between \answer{<answer>} and \answer{</answer>}. If it chooses to invoke the image search tool, it should append \search{<search><img></search>} at the end of its response. For text search, the model should autonomously generate a natural language query and append it between \search{<text\_search>} and \search{</text\_search>}. All retrieved information is returned to the model enclosed within \information{<information>} and \information{</information>}, and is fed into the next round of dialogue. This iterative process continues until the model provides a final answer or reaches the maximum number of allowed turns. To prevent training bias from environment feedback, retrieved content from search tools is masked during loss computation and does not contribute to gradient updates.

\subsection{Reward Modeling}
Reward modeling plays a critical role in RL training, as it encodes the desired model behavior and directly guides the optimization process. In MMSearch-R1, the reward consists of two components: an accuracy score with search penalty and a format score. 

\textbf{Accuracy Score with Search Penalty} We use the exact string match to evaluate whether the final answer provided by the model is consistent with the ground truth. If the match is exact, the accuracy score is 1; otherwise, it is 0. For correct answers, we further check whether the model relied on search tools to arrive at the answer. A penalty factor (ranging from 0 to 1) is applied to the accuracy score when any search was performed. This design encourages the model to first exploit its internal knowledge and invoke search tools only when necessary, thereby shaping on-demand search behavior.

\textbf{Format Score} This component checks whether the multi-turn responses of the model strictly follow the predefined prompt format. Specifically, the model is required to reason between \reason{<reason>} and \reason{</reason>} before taking any action, take only one action per turn, place search patterns such as \search{<search><img></search>} or \search{<text\_search> text query </text\_search>} at the end of its response when a search is performed, and wrap the final answer between \answer{<answer>} and \answer{</answer>} when concluding. The format score is assigned a value of 1 only if all responses fully adhere to these formatting requirements; otherwise, it is set to 0. We combine the two components using a weighting coefficient $\alpha$, and define the final reward as a weighted sum of the accuracy and format scores as follows:
{\normalsize
\begin{equation}
    reward = (1 - \alpha)\cdot Acc\_Score\cdot Search\_Penalty + \alpha\cdot Format\_Score.
\end{equation}
}
\section{Curating Search-balanced VQA Datasets}
\label{method:data}

\begin{figure}[t]
  \centering
  \includegraphics[width=0.96\textwidth]{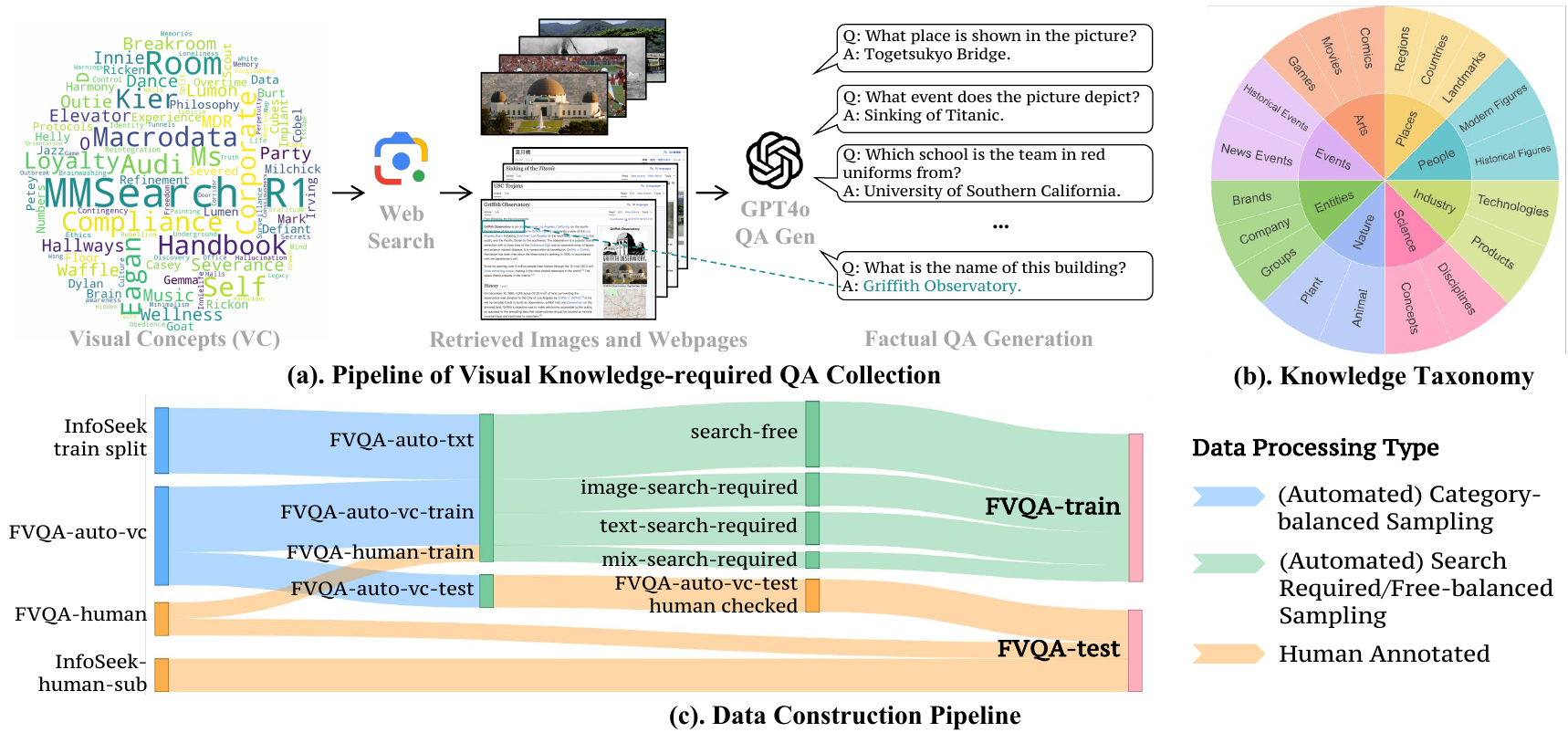}
  \caption{Illustration of data construction process of \texttt{FVQA}: (a). An automated pipeline for visual knowledge-required VQA samples collection; (b). Knowledge taxonomy; (c). Overall pipeline showing the composition and origin of \texttt{FVQA} from various automatic and manually curated sources.}
  \label{fig:fvqa_data_pipeline}
\end{figure}

We aim to explore the potential of training models for on-demand search using simple outcome-based reward reinforcement learning. Therefore, the datasets should satisfy three criteria: \textbf{(1). Coverage of Both Search-Required and Search-Free Questions} The datasets should include a mix of search-free and search-required questions. A search-free question can be answered solely using the model’s internal knowledge, whereas a search-required question involves information beyond the model’s existing knowledge and therefore requires access to external information sources. Search-required questions can be further categorized into Visual Knowledge-required and Textual Knowledge-required types. Visual Knowledge refers to the model’s ability to recognize visual entities in an image (e.g., "What is the model of the aircraft shown in the image?"), while Textual Knowledge refers to factual information about the visual entity (e.g., "Who is the designer of Supermarine Spitfire?"). Existing datasets such as OK-VQA \cite{dataset/ok-vqa} and InfoSeek \cite{dataset/infoseek} typically use image sources already seen by current multimodal models and tend to focus on Textual Knowledge-required questions. Meanwhile, MMSearch \cite{vision-rag/mmsearch} collects images from recent news, aligning better with the Visual Knowledge-required criteria. However, it is relatively small in scale and mainly serve as benchmarks, leading to a lack of sufficient Visual Knowledge-required training data. \textbf{(2). Concise and Unambiguous Answers for Reliable Verification} VQA questions should be designed to elicit concise and unambiguous answers that can be easily and reliably verified through simple, rule-based reward mechanisms. Such questions typically revolve around factual knowledge, enabling automated evaluation during reinforcement learning. \textbf{(3). Diversity in Knowledge Categories and Question Difficulty} The dataset should cover diverse knowledge domains and difficulty levels to ensure broad real-world generalization. 

To support our investigation, we construct a multimodal search VQA dataset, \texttt{FactualVQA (FVQA)}, using a combination of automated pipelines and manual annotation, covering both training and evaluation needs, as illustrated in Figure~\ref{fig:fvqa_data_pipeline}(c).

\subsection{Training Dataset Construction}
To meet the above requirements, we select suitable training data from multiple sources through two main processes: VQA Collection and Search Balancing.

\textbf{VQA Collection} We first developed an automated annotation pipeline to collect Visual Knowledge-required training data, as illustrated in Figure~\ref{fig:fvqa_data_pipeline}(a). To explore visual concepts that models are relatively familiar or unfamiliar with, we begin with the Metadata distribution of MetaCLIP \cite{dataset/metaclip}. This Metadata is constructed from multiple sources such as WordNet and Wikipedia, covering a wide range of Visual Concepts (VC) from common to rare. Designed to guide balanced dataset construction for vision-language pretraining, it exhibits a long-tailed distribution. Concepts in the head of the distribution correspond to commonly referenced items in the real world (e.g., car, tree), whereas those in the tail represent less common or niche concepts (e.g., thylacine, astrolabe). We randomly sampled 10,000 visual concepts from both the head and the tail of the Metadata distribution. For each concept, we performed a web search to retrieve the most relevant image and its associated webpage. These image–webpage pairs were then input to GPT-4o, which was prompted to generate a factual visual question-answer (VQA) pair centered around the given visual concept. The generated answers were required to be concise and precise.
The prompts used for QA generation and two examples can be found in Appendix~\ref{appendix:train-data-gen-prompt} and~\ref{appendix:example-fvqa}.
Next, GPT-4o was employed to classify the knowledge type assessed by each question, resulting in a knowledge taxonomy as illustrated in Figure~\ref{fig:fvqa_data_pipeline}(b). Based on this taxonomy, we performed balanced sampling across categories and curated a final set of 6,000 VQA samples, referred to as \texttt{FVQA-auto-vc}, with 5,400 used for training and 600 for testing. In parallel, to enrich the dataset with textual knowledge-required examples, we sampled from the open-source InfoSeek \cite{dataset/infoseek} dataset. Specifically, we classified the questions in the InfoSeek training split by the type of knowledge they require and mapped them into the same taxonomy. After balanced sampling across categories, we obtained 7,000 samples, referred to as \texttt{FVQA-auto-txt}. Finally, to further diversify the dataset with real-user queries, we manually annotated an additional 800 samples, referred to as \texttt{FVQA-manual-train}. Annotators were instructed to select a knowledge category from the taxonomy, locate a relevant image, and pose a factual question pertaining to that category. They were allowed to perform both visual and textual searches until sufficient information was gathered to formulate an accurate answer, from which a concise and precise response was extracted.

\textbf{Search Balancing} The goal of this stage is to distinguish between search-required and search-free questions within the collected data. To this end, we first trained a Qwen2.5-VL-Instruct-7B model using our training framework on the full set of samples obtained from the VQA collection process. This model was then used to classify the original questions: for each question, 8 rollouts were generated. If all 8 rollouts failed, the question was discarded due to insufficient training signal. A question was labeled as image-, text-, or mixed-search-required if the model produced correct answers only when the corresponding type of search behavior was performed. In particular, mixed-search indicates that both image and text searches needed to be executed for the model to answer correctly. If any rollout produced a correct answer without invoking search, the question was labeled as search-free. Finally, we constructed a search-balanced training set of 5,000 samples, referred to as \texttt{FVQA-train}, consisting of approximately 3,400 search-required and 1,600 search-free VQA examples. Maintaining a balanced distribution of search types is crucial to shaping the search behavior of the model during training.

\subsection{Test Dataset Annotation}
To better evaluate the model’s performance, we additionally constructed a high-quality test set, where all examples were either manually verified or fully human-annotated to ensure accuracy. The test set, referred to as \texttt{FVQA-test}, includes 1800 examples collected from three sources: (1) 600 samples drawn from \texttt{FVQA-auto-vc}, ensuring no overlap with the training set, with each example manually checked for correctness; (2) 600 samples selected from the InfoSeek Human Split, where answers were manually annotated, as the original human-labeled answers were not publicly available; (3) 600 samples collected directly from the manual annotation process described above.
\section{Experiments}

\begin{table}[t]
\footnotesize
\centering
\caption{Performance of MMSearch-R1 across benchmarks. 
"Acc (\%)" denotes the accuracy evaluated by LLM-as-Judge, while "SR (\%)" represents the search ratio, defined as the percentage of total search calls made relative to the maximum allowed search steps for each method.}
\label{tab:finding-1}
\setlength{\tabcolsep}{4.5pt}
\begin{tabular}{ccccccccccccc}
\toprule
\multicolumn{1}{c|}{\multirow{3}{*}{\textbf{Model}}} & \multicolumn{2}{c|}{\multirow{2}{*}{\textbf{Average}}} & \multicolumn{4}{c|}{\textbf{In-Domain}} & \multicolumn{6}{c}{\textbf{Out-of-Domain}} \\
\multicolumn{1}{c|}{} & \multicolumn{2}{c|}{} & \multicolumn{2}{c}{\textbf{FVQA-test}} & \multicolumn{2}{c|}{\textbf{InfoSeek}} & \multicolumn{2}{c}{\textbf{MMSearch}} & \multicolumn{2}{c}{\textbf{SimpleVQA}} & \multicolumn{2}{l}{\textbf{LiveVQA}} \\
\multicolumn{1}{c|}{} & \textbf{Acc} & \multicolumn{1}{c|}{\textbf{SR}} & \textbf{Acc} & \textbf{SR} & \textbf{Acc} & \multicolumn{1}{c|}{\textbf{SR}} & \textbf{Acc} & \textbf{SR} & \textbf{Acc} & \textbf{SR} & \multicolumn{1}{c}{\textbf{Acc}} & \multicolumn{1}{c}{\textbf{SR}} \\ \midrule
\multicolumn{13}{c}{\textit{\textbf{Direct Answer}}} \\ \midrule
\multicolumn{1}{c|}{GPT4o} & 36.0 & \multicolumn{1}{c|}{0} & 41.7 & 0 & 42.7 & \multicolumn{1}{c|}{0} & 22.2 & 0 & 46.6 & 0 & 26.9 & 0 \\
\multicolumn{1}{c|}{Gemini 2.5 Pro} & 36.4 & \multicolumn{1}{c|}{0} & 37.2 & 0 & 37.0 & \multicolumn{1}{c|}{0} & 26.9 & 0 & 53.4 & 0 & 27.7 & 0 \\
\multicolumn{1}{c|}{Qwen2.5-VL-72B} & 26.6 & \multicolumn{1}{c|}{0} & 27.1 & 0 & 28.0 & \multicolumn{1}{c|}{0} & 15.7 & 0 & 42.2 & 0 & 20.1 & 0 \\
\multicolumn{1}{c|}{Qwen2.5-VL-32B} & 25.0 & \multicolumn{1}{c|}{0} & 24.7 & 0 & 25.8 & \multicolumn{1}{c|}{0} & 15.7 & 0 & 40.1 & 0 & 18.7 & 0 \\
\multicolumn{1}{c|}{Qwen2.5-VL-7B} & 21.9 & \multicolumn{1}{c|}{0} & 20.3 & 0 & 20.1 & \multicolumn{1}{c|}{0} & 12.8 & 0 & 38.4 & 0 & 17.8 & 0 \\ \midrule
\multicolumn{13}{c}{\textit{\textbf{RAG Workflow}}} \\ \midrule
\multicolumn{1}{c|}{GPT4o} & 62.1 & \multicolumn{1}{c|}{100} & 66.0 & 100 & 59.1 & \multicolumn{1}{c|}{100} & 62.5 & 100 & 63.4 & 100 & 59.6 & 100 \\
\multicolumn{1}{c|}{Gemini 2.5 Pro} & 61.8 & \multicolumn{1}{c|}{100} & 66.1 & 100 & 56.7 & \multicolumn{1}{c|}{100} & 62.5 & 100 & 65.9 & 100 & 57.8 & 100 \\
\multicolumn{1}{c|}{Qwen2.5-VL-72B} & 59.6 & \multicolumn{1}{c|}{100} & 62.2 & 100 & 59.4 & \multicolumn{1}{c|}{100} & 59.6 & 100 & 61.0 & 100 & 56.0 & 100 \\
\multicolumn{1}{c|}{Qwen2.5-VL-32B} & 55.1 & \multicolumn{1}{c|}{100} & 57.0 & 100 & 56.8 & \multicolumn{1}{c|}{100} & 57.9 & 100 & 54.5 & 100 & 49.6 & 100 \\
\multicolumn{1}{c|}{Qwen2.5-VL-7B} & 51.6 & \multicolumn{1}{c|}{100} & 52.9 & 100 & 53.7 & \multicolumn{1}{c|}{100} & 52.2 & 100 & 51.6 & 100 & 48.0 & 100 \\ \midrule
\multicolumn{13}{c}{\textit{\textbf{On-demand Search}}} \\ \midrule
\rowcolor{lightgreen}
\multicolumn{1}{c|}{\textbf{MMSearch-R1-7B}} & 54.6 & \multicolumn{1}{c|}{67.1} & 58.4 & 66.8 & 55.1 & \multicolumn{1}{c|}{61.6} & 53.8 & 88.5 & 57.4 & 42.5 & 48.4 & 76.2 \\ \bottomrule
\end{tabular}
\end{table}

\subsection{Setups}
\label{main/exp-setups}
\textbf{Implementation Details} We built our training framework based on veRL \cite{code/verl} and conduct experiments on Qwen2.5-VL-7B-Instruct \cite{lmm/qw25vl}. The model was trained using the dataset described in Section~\ref{method:data}. At each training step, we sampled 512 examples, with each example undergoing 8 rollouts. Each rollout consists of up to three rounds of dialogue, during which the model can perform at most two search actions and is required to produce a final answer in the third round. Image search is only allowed in the first round, and each image search returns up to 5 top visual matched webpages in the form of interleaved thumbnails and titles. Text search, on the other hand, returns up to 5 summarized webpage contents per query. The search penalty factor is set to 0.1. The weighting coefficient $\alpha$ between the accuracy reward and the format reward is set to 0.1. Further details on the hyperparameter settings can be found in Appendix~\ref{appendix:exp-training-hyperparameters-rl}.

\textbf{Benchmark} We selected \texttt{FVQA-test}, InfoSeek \cite{dataset/infoseek}, MMSearch \cite{vision-rag/mmsearch}, SimpleVQA \cite{dataset/simplevqa} and LiveVQA \cite{dataset/livevqa} as benchmark datasets to evaluate the model's ability to handle both knowledge-intensive and information-seeking VQA tasks. Specifically, for InfoSeek, we randomly sampled 2k examples from its test split due to the large dataset size. For MMSearch, we filter and retain only the QA pairs that include images. For SimpleVQA, we extracted all QA examples written in English. Among these benchmarks, MMSearch, SimpleVQA and LiveVQA serve as out-of-distribution (OOD) testsets for our trained models. Details of the benchmark datasets are provided in Appendix~\ref{appendix:benchmark-details}.

\textbf{Baselines} To validate the effectiveness of MMSearch-R1, we evaluated against both closed-source models (GPT-4o and Gemini 2.5 Pro) and open-source models from the Qwen2.5-VL series. All models are tasked with solving VQA problems in two different workflows. (1) Direct Answer: Models are prompted to directly generate a short and precise answer without accessing external information. (2) Answer under RAG Workflow: In this workflow, models are required to perform exactly two search operations using our multimodal search tools for each VQA example, first performing an image search and then a text search. Specifically, given an input image and question, the model is provided with the image search results and the original question in the first round and is prompted to generate a text query to assist in answering. In the second round, the retrieved results based on the text query are fed into the model, and the model is asked to produce the final answer. Under a fixed budget of search steps, the RAG workflow typically exposes the model to more external information compared to the on-demand search strategy. The prompts used for the RAG workflow are provided in Appendix~\ref{appendix:rag-prompts}.

\textbf{Metric} We adopt the LLM-as-Judge framework as our evaluation metric to assess the accuracy of model responses. In addition, we record the search ratio, the proportion of responses that invoke a search, for each method across different benchmarks, allowing us to jointly evaluate answer correctness and search efficiency. Specifically, we employ GPT-4o as the judging model, which receives the original image, the question, the ground-truth answer, and the model’s response to determine correctness. The prompt used for judgment is detailed in the Appendix~\ref{appendix:eval-prompts}.

\subsection{Findings}
In this section, we present key empirical findings that emerged from our experiments. Each finding is supported by quantitative results and detailed analysis, aiming to provide deeper insights into the behavior and effectiveness of MMSearch-R1. Through these findings, we aim to both validate our approach and provide actionable insights for the broader research community.

\paragraph{\textit{Finding 1: RL training enables models to better recognize the boundaries of their knowledge and perform on-demand search more effectively.}} As shown in Table~\ref{tab:finding-1}, on both in-domain and out-of-domain test sets, MMSearch-R1-7B outperforms the RAG-based counterparts of the same size by an average of \textbf{3\%} in accuracy, while reducing the average search rate by \textbf{32.9\%}. This indicates that our RL-trained model not only achieves higher correctness but also relies less on external information, thereby exhibiting a more efficient and targeted use of search. Notably, MMSearch-R1-7B performs competitively with RAG-based Qwen2.5-VL-32B, a significantly larger model, further highlighting the benefits of learning adaptive search behavior, as opposed to executing fixed-stage retrieval regardless of necessity. These results validate the effectiveness of outcome-based RL in enabling intelligent, cost-aware search behavior.

\begin{figure}[t]
  \centering
  \includegraphics[width=0.99\textwidth]{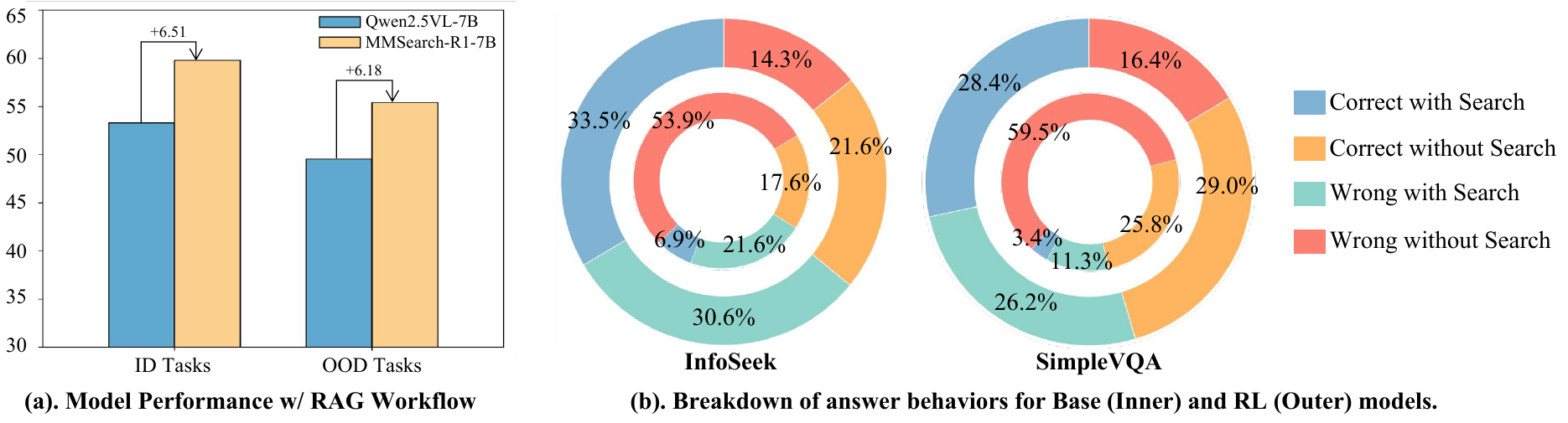}
  \caption{(a). Performance comparison between the Base model and the RL-trained model under the RAG workflow. (b). Answer behavior breakdown of Base (inner circle) and RL (outer circle) models in InfoSeek and SimpleVQA.}
  \label{fig:finding_2_3}
\end{figure}

\paragraph{\textit{Finding 2: RL training enhances the model’s ability to generate effective text queries and summarize retrieved information.}} 
Our RL training jointly equips model with three key capabilities: deciding whether to search, determining what to search for, and extracting useful information from retrieved results. To evaluate the latter two abilities in isolation, we adopt the same RAG setting as in the baseline comparison, where both image and text search are executed for every question. In this setup, the search process is fixed, and the model is responsible only for generating effective text queries and reasoning over the retrieved content. This removes variability in search triggering and allows us to focus on evaluating how well the model interacts with external information.

As shown in Figure~\ref{fig:finding_2_3}(a), MMSearch-R1-7B demonstrates consistent improvements over the base model across both in-domain and out-of-domain tasks. These results indicate that reinforcement learning improves not only the decision of when to search, as shown earlier, but also enhances two core retrieval abilities: generating more accurate, contextually relevant queries and extracting useful information to support accurate answers. These gains appear across both image and text retrieval, highlighting the broader value of RL in strengthening retrieval and reasoning capabilities.

\paragraph{\textit{Finding 3: RL improves the model's ability to utilize its internal knowledge.}} As shown in Figure~\ref{fig:finding_2_3}(b), we conduct a behavioral breakdown of responses on datasets InfoSeek and SimpleVQA to better understand how the model’s behavior changes after reinforcement learning. The results reveals that a clear upward trend in the \textit{Correct without Search} proportion from the base model to the RL-trained model. These gains indicate that the RL-trained model can answer substantially more questions correctly without invoking the search tool, demonstrating improved recall and reasoning based on its internal knowledge. This shift suggests that reinforcement learning enhances the model's ability to rely on its own parameters when sufficient, and to reserve external search for genuinely novel or long-tail queries. As a result, the model exhibits more accurate on-demand search behavior, engaging external tools only when internal knowledge is insufficient.

\paragraph{\textit{Finding 4: RL achieves greater performance improvements and exhibits higher data efficiency compared to supervised SFT.}} To investigate the effectiveness of RL and SFT as training paradigms, we compare their performance gains and data efficiency through controlled experiments. For RL, we use the aforementioned \texttt{FVQA-train} set, which contains 5k VQA samples in the form of \texttt{(image, question, answer)} triplets. For SFT, we first construct a dataset by aggregating three sources: \texttt{FVQA-auto-vc}, \texttt{FVQA-auto-txt}, and \texttt{FVQA-human-train}, resulting in 13.2k examples. We then distill GPT-4o's behavior on this dataset by prompting it with the same instructions used in RL training, allowing it to autonomously decide whether to invoke external tools when answering a question. This process yields 13.2k responses from GPT-4o, each including an image, question, and a response containing its reasoning process and any tool usage, often in a multi-turn dialogue format. During SFT training, we mask out the prompts and all tool-returned content from the multi-turn dialogues, using only GPT-4o's responses as supervision signals. This ensures that the model learns to mimic GPT-4o’s reasoning and answering behavior without relying on intermediate tool outputs, thereby encouraging it to internalize the reasoning process demonstrated by the teacher model. To ensure training quality, we compare GPT-4o's final predicted answers with the original ground truth answers and filter out 5.2k samples where GPT-4o produced incorrect answers. The remaining 8k examples constitute our SFT training set. The implementation details of SFT can be found in the Appendix~\ref{appendix:exp-training-hyperparameters-sft}.

\begin{figure}[t]
  \centering
  \includegraphics[width=0.99\textwidth]{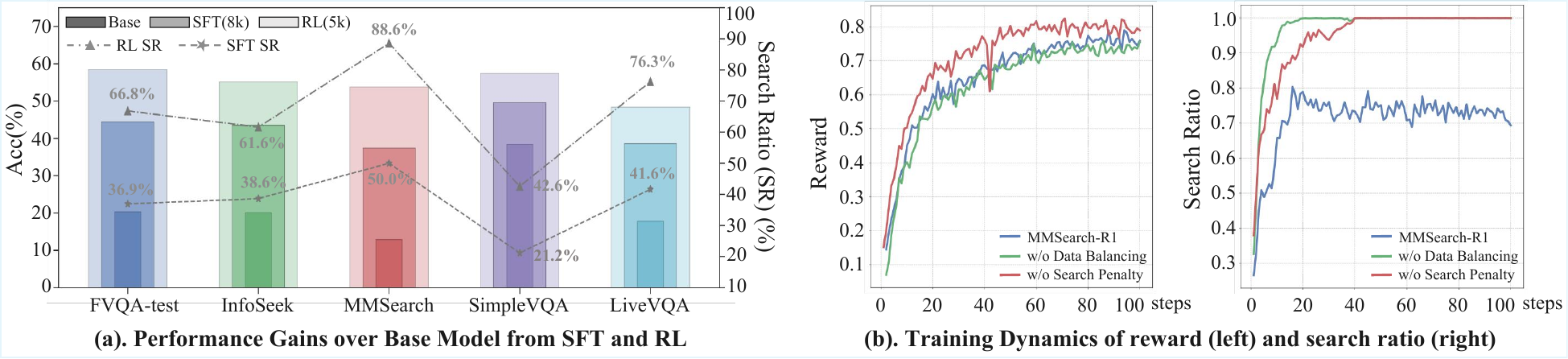}
  \caption{(a). Performance improvements of SFT and RL over Base across five VQA datasets. (b). Training dynamics of reward and search ratio for different strategies}
  \label{fig:finding_4_5}
\end{figure}

We fine-tune the base model, Qwen2.5-VL-7B, using the above datasets for both SFT and RL training. We then evaluate the performance of the models trained with SFT and RL on all downstream tasks and compare their improvements over the base model, as shown in Figure~\ref{fig:finding_4_5}(a). The results show that the model trained with RL consistently outperforms the one trained with SFT across all tasks, despite being trained on only about half as much data. In addition, the behavior of the RL-trained model aligns more closely with the specific requirements of each task, especially in its use of search tools. For example, in the MMSearch and LiveVQA tasks, which are both recently collected and focus on information-seeking questions, the RL-trained model shows a higher frequency of search tool invocation. This is consistent with expectations, as these datasets contain many questions that require external visual or textual knowledge not encountered during pretraining. The RL-trained model learns to rely more effectively on search tools in such scenarios.

\paragraph{\textit{Finding 5: Training with balanced data and a search penalty in the reward effectively guide the model to perform on-demand search.}} Figure~\ref{fig:finding_4_5}(b) illustrates the training dynamics of reward and search ratio during reinforcement learning. Removing either the search penalty or data balancing leads to distinct trade-offs. Although both ablated variants achieve slightly higher rewards, they do so at the cost of overusing the search tool, with search ratios rapidly converging to nearly 100\%. In contrast, MMSearch-R1, trained with both data balancing and a search penalty, achieves a comparable reward while maintaining a significantly lower and more stable search ratio. This suggests that the model has learned to invoke the search tool only when necessary, enabling more efficient and selective on-demand search behavior.
\section{Conclusion}
In this work, we present MMSearch-R1, a RL-based framework that equips LMMs with the ability to perform on-demand search in real-world internet environments. MMSearch-R1 learns to recognize knowledge gaps, selectively invoke image or text search, and reason over retrieved content. It outperforms same-sized RAG baselines and approaches the performance of larger models while requiring significantly fewer search calls. Our framework, dataset, and findings offer practical insights into training LMMs with real-world interaction capabilities and lay the groundwork for building multimodal agents that are both adaptive and interactive. We look forward to the next major advancement in multimodal intelligence emerging as models increasingly engage with and explore the real world through more tools, further evolving their reasoning and adaptive capabilities.


\newpage
\bibliographystyle{plain}
\bibliography{main}

\newpage
\appendix
\section{Related Work}
\subsection{Large Multimodal Models (LMMs)}
The development of Large Multimodal Models (LMMs) marks a significant breakthrough in artificial intelligence, enabling unified processing of textual and visual information that aligns with the inherently multimodal nature of human perception. Recent advances have established a dominant paradigm where scaling up diverse, high-quality vision-text paired datasets across training stages equips LMMs with grounded visual understanding. This approach has yielded remarkable success, as demonstrated by state-of-the-art models including GPT-4o~\cite{lmm/gpt4o}, Gemini~\cite{lmm/gemini}, Qwen2.5-VL~\cite{lmm/qw25vl}, the LLaVA series~\cite{lmm/llava,lmm/llava-ov,lmm/llava-video}, and related systems~\cite{lmm/claude-3.5,lmm/llama-3.2,lmm/aira,lmm/internvl,lmm/vila}, which exhibit exceptional capabilities in visual comprehension, cross-modal reasoning, and instruction following. However, this training methodology inherently yields static knowledge, while real-world information is complex, dynamic, and constantly evolving. As a result, LMMs often struggle with long-tail concepts or newly emerging facts beyond their training cut-off, which can lead to hallucinations and ultimately compromise their stability and reliability in real-world applications. This limitation has led to increasing efforts to augment LMMs with retrieval and search tool-use capabilities.

\subsection{Large Models with External Knowledge Access}
Given the strong reliance of large models on external knowledge in practical applications, two primary approaches have emerged to enhance their factual reliability and knowledge coverage: retrieval-based methods and search-based methods. 

\subsubsection{Retrieval-Augmented Generation (RAG)}

The RAG paradigm retrieves relevant information from external knowledge bases via dense vector search and incorporates it into the model input to help generate more factually grounded responses. In the field of natural language understanding, Petr Karpukhin et al.~\cite{text-rag/DPR} introduced Dense Passage Retrieval (DPR), which was the first to apply dense retrieval to open-domain question answering. DPR employs a dual-encoder architecture to separately encode queries and document passages, enabling efficient semantic-level matching. Building on DPR’s strong retrieval capabilities, Lewis et al.~\cite{text-rag/RAG} proposed the RAG framework, which integrates pre-trained generative models with nonparametric document indices, forming a unified 'retrieval-then-generation' architecture that laid the foundation for subsequent research. As model training and inference techniques have advanced, RAG-based methods have gradually incorporated key techniques such as long-context modeling, chain-of-thought reasoning, self-reflection and document ranking. These enhancements have improved performance across pretraining~\cite{text-rag/instrcutretro-pretrain}, fine-tuning~\cite{text-rag/atlas-finetune,text-rag/raft-finetune,text-rag/uprise-finetune}, and inference~\cite{text-rag/itergen-infer,text-rag/selfrag-inference,text-rag/RankRAG-infer,text-rag/MDR-infer} stages, significantly boosting RAG’s effectiveness in knowledge-intensive tasks. 

The field of visual understanding has also seen notable progress through the adoption of retrieval-augmented methods, particularly in knowledge-intensive VQA tasks. By incorporating external knowledge, including text, images, and structured data, these approaches overcome the limitations of vision-only models that often lack access to broader world knowledge. MuRAG~\cite{vision-rag/murag} integrates both image and text retrieval to enhance open-domain question answering over multimodal data. REVEAL~\cite{vision-rag/reveal} utilizes a multi-source multimodal knowledge memory to augment visual-language pre-training, enabling better handling of knowledge-intensive queries. RagVL~\cite{vision-rag/rag-vl} introduces a knowledge-enhanced reranking mechanism, improving retrieval precision by filtering out noisy data. VisRAG~\cite{vision-rag/visrag} employs a vision-based approach to process multi-modality documents, preserving visual information without relying solely on text extraction. Collectively, these advances have demonstrated the potential of retrieval-augmented generation to bridge visual understanding with external knowledge.

Although RAG methods have shown strong performance in practice, they face several critical limitations, the most notable being their reliance on a static knowledge base. These methods often assume that the required information can be found within the existing corpus. However, in real-world scenarios, information is frequently dynamic and constantly evolving, and the complexity of web-scale environments means that relevant content may not always be effectively retrieved. This poses significant challenges to the applicability of RAG-based approaches in more general and open-ended settings.

\subsubsection{Search-Augmented Approaches}

Several recent efforts have explored search-augmented paradigms that go beyond static corpora by enabling models to interact with real-time information sources. In the field of natural language understanding, WebGPT~\cite{text-search/webgpt} is one of the earliest systems to demonstrate this idea at scale. It augments a language model with access to Bing search results and trains it using human feedback to quote sources and generate more factual, citeable answers. This setup significantly improves factual consistency, especially in domains requiring timely or less common knowledge. Toolformer~\cite{text-search/toolformer} introduces a self-supervised framework where the model learns when and how to invoke external tools, such as search engines, by generating API call demonstrations and fine-tuning on helpful examples. This enables efficient and context-aware tool use without heavy supervision. SAIL~\cite{text-search/sail} further integrates web search into instruction tuning. It constructs training examples by pairing user prompts with retrieved search results, teaching the model to select relevant information, filter noise, and perform multi-hop reasoning. While these approaches demonstrate strong potential for dynamic and tool-augmented reasoning, they still rely heavily on high-quality annotated data for training. As language models continue to improve in both knowledge retention and reasoning capabilities, a new line of training-free, prompt-based methods has emerged. These approaches leverage large models as agents orchestrated by structured prompts, without requiring additional fine-tuning. For example, Recent studies~\cite{text-search/deepresearcher,text-search/open-deep-search} have introduced agentic workflows that incorporate web search tools directly into the reasoning process. By decomposing complex queries into sub-tasks and selectively invoking external tools through prompts, these systems can effectively tackle information-seeking tasks that require up-to-date and reliable web content. Such approaches highlight the growing viability of zero-shot and in-context tool use for knowledge-intensive applications. 

Building on this trend, recent research has further extended tool-augmented, training-free paradigms into the multimodal domain, exploring how large models can function as autonomous visual search agents. For example, AVIS~\cite{vision-rag/avis} proposes a framework in which a large language model acts as a controller, iteratively issuing vision-language queries to retrieve external visual and textual information in order to answer complex visual questions. VSA~\cite{vision-rag/vsa} takes this further by equipping vision-language models with retrieval capabilities, enabling them to behave like multimodal search engines that can proactively seek and integrate visual evidence from large-scale corpora. MMSearch~\cite{vision-rag/mmsearch} presents a comprehensive pipeline that empowers large multimodal models with advanced search capabilities. Its agentic workflow encompasses re-querying, reranking, and summarization stages, enabling models to autonomously process and synthesize information from both visual and textual modalities. These approaches signify a shift towards interactive, tool-augmented processes in multimodal reasoning. However, a key limitation of these approaches is that the underlying models are typically not trained to interact with search tools in a supervised or reinforcement-based manner. As a result, the agent may not learn to engage with external tools in the most effective or reliable way, especially in real-world environments with noisy or ambiguous search results.

\subsection{Reinforcement Learning-powered Search Agents}
Recent advances, such as OpenAI's o series~\cite{rl/openai-o1,openai-o3_o4_mini_syscard}, DeepSeek-R1~\cite{rl/deepseek-r1} and Kimi-K1.5~\cite{rl/kimi-k1.5}, have highlighted the potential of end-to-end reinforcement learning (RL) to enhance the reasoning capabilities of large-scale models. These efforts have led to the emergence of Large Reasoning Models (LRMs) that are capable of solving complex, multi-step reasoning tasks beyond the reach of standard instruction-tuned language models. Building on this momentum, leading organizations such as OpenAI, Google, and Perplexity have introduced Deep Research agents~\cite{openai-deep-research,claude-deep-research,gemini-deep-research,perplexity-deep-research}. These systems combine large reasoning models with real-time web search tools to complete open-ended, research-oriented tasks, such as drafting reports or synthesizing diverse information sources. Notably, OpenAI reports that it has successfully trained a highly capable Deep Research model through end-to-end reinforcement learning, demonstrating the feasibility and effectiveness of this approach in real-world applications. In the open-source community, efforts such as DeepResearcher \cite{text-search/deepresearcher}, Search-R1 \cite{text-search/search-r1}, and ReSearch \cite{text-search/ReSearch} have followed this direction, applying end-to-end RL to improve models’ abilities in multiturn search and retrieval-augmented generation, aiming to boost performance on information-seeking question answering tasks. However, existing work mainly focuses on text-based search, while in the multimodal domain, challenges such as constructing suitable training data and designing RL frameworks to incentivize models to perform search in real-world environments remain underexplored.

\section{Group Relative Policy Optimization (GRPO) Algorithm}
\label{appendix:grpo}
GRPO updates the current policy model using a group of rollouts sampled from an old policy $\pi_{old}$, along with a reference $\pi_{ref}$ used as a constraint. Specifically, given a question $q$ sampled from the training set $\mathcal{D}$, GRPO generates a set of outputs from $\pi_{old}$, and optimizes the current policy $\pi_{\theta}$ by maximizing the following objective:
\begin{equation}
\begin{aligned}
\mathcal{J}_{GRPO}(\theta) & =\mathbb{E}[q\sim \mathcal{D},\{o_{i}\}_{i=1}^{G}\sim\pi_{\theta_{old}}(O|q)] \\
 & \frac{1}{G}\sum_{i=1}^{G}\frac{1}{|o_{i}|}\sum_{t=1}^{|o_{i}|}\left\{\min\left[R_{i,t}\hat{A}_{i,t}\operatorname{clip}\left(R_{i,t},1-\varepsilon,1+\varepsilon\right)\hat{A}_{i,t}\right]-\beta\mathbb{D}_{KL}\left[\pi_{\theta}\|\pi_{ref}\right]\right\},
\end{aligned}
\end{equation}
where $R_{i,t}=\frac{\pi_{\theta}(o_{i,t}|q,o_{i,<t})}{\pi_{\theta_{old}}(o_{i,t}|q,o_{i,<t})}$ is the policy ratio. $\epsilon$ and $\beta$ are hyper-parameters that control the stability and constraint strength of the policy update, respectively. Given the sampled rollouts $\mathbf{o} = \{o_1, o_2, ..., o_G\}$, a group reward $\mathbf{r} = \{r_1, r_2, \cdots, r_G \}$ can be obtained by a reward model. The advantage $\hat{A}_{i,t} = \tilde{r}_i = \frac{r_i - \operatorname{mean}(\mathbf{r})}{\operatorname{std}(\mathbf{r})}$ mitigates variance across samples and prevents any single reward signal from dominating the policy update.

\section{Multimodal Search Tools}
\label{appendix:mm-search-tools}

\begin{figure}[t]
  \centering
  \includegraphics[width=0.99\textwidth]{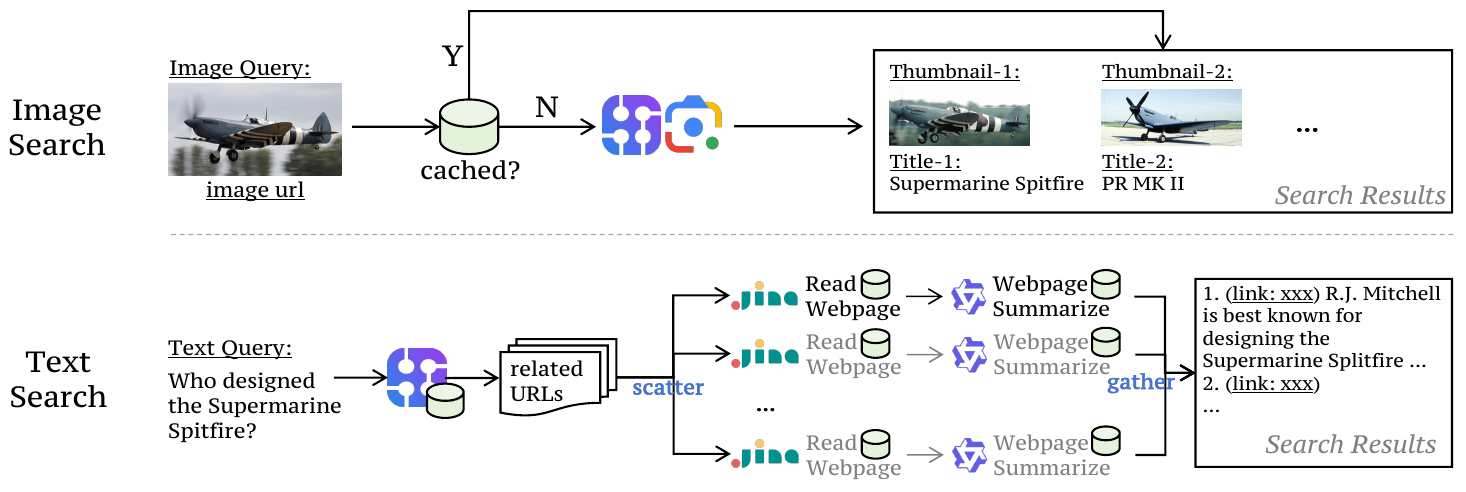}
  \caption{The overall architecture of the multimodal search pipeline.}
  \label{fig:apd_mm_search_tools}
\end{figure}

To enhance the stability and efficiency of multimodal search services during both training and inference, we encapsulate the image and text search tools as independent HTTP services. These services are optimized through pipelined parallel processing, local caching, and multi-metric monitoring. The overall architecture of the search pipeline is illustrated in Figure~\ref{fig:apd_mm_search_tools}.

\textbf{Image Search Tool} The image search tool is built on top of SerpAPI. Given an input image URL, SerpAPI returns a set of visually similar webpages, including metadata such as URLs, thumbnails, and titles. We sort the returned results by relevance and extract up to five valid results (thumbnail + title pairs). To avoid redundant SerpAPI calls during training and evaluation, we implement a cache mapping image URLs to their corresponding search results.

\textbf{Text Search Tool} The text search pipeline consists of SerpAPI, JINA Reader, and a webpage summarization model, forming a search–parse–summarize chain. SerpAPI first retrieves a list of webpage URLs based on a given query. In each round, the top 5 URLs are processed using a pipelined parallel strategy: JINA fetches and converts the webpage content into clean, structured text, while Qwen3-32B acts as a summarizer, focusing on extracting information relevant to the original user question using a tailored prompt. The full prompt is provided in Tabel~\ref{tab:webpage-summary-prompt}. This approach improves focus and reduces token usage. If some URLs fail to be processed, additional URLs from the SerpAPI result list are used to retry, until 5 successful summarizations are obtained or the list is exhausted.To improve efficiency and reduce redundant computation, the text search tool incorporates a multi-layer caching mechanism: (1). Mapping from query to webpage URLs (to skip repeated searches); (2). Mapping from URL to JINA parsing results (to avoid re-parsing); (3). Mapping from JINA outputs to Qwen3-32B summaries (to reduce summarization cost). We deployed Qwen3-32B on a cluster of 8 $\times$ 8 NVIDIA H100 GPUs to support our training and inference needs.

\textbf{Cache Mechanism} Caching is implemented using Redis to support high-throughput concurrent read/write access. For latency-critical components, caching is applied to reduce response time. Because JINA parsing results are often large, they are stored in object storage (TOS), while Redis only stores references (keys) to these results, reducing memory footprint. We adopt an LRU (Least Recently Used) policy for eviction under memory constraints. Additionally, a scheduled offline retry mechanism is in place to reprocess failed JINA URLs and update the cache accordingly.

\textbf{Distributed Rate Limiting} infrastructure-level constraints on concurrent request handling can bottleneck training efficiency. Limited parallelism may cause request backlogs or latency spikes, ultimately affecting throughput and stability. To manage system load during peak traffic, we implement distributed rate limiting using Redis. This allows us to smooth bursty traffic to JINA and webpage summary model over time windows, reducing the risk of system overload and improving service availability and stability.

\section{Prompts}
\subsection{Prompts for FVQA-auto-ac VQA Generation}
\label{appendix:train-data-gen-prompt}
To generate factual QA pairs for the \texttt{FVQA-auto-vc} set, we designed the prompt to guide GPT-4o in producing concise and grounded question–answer pairs based on given image–webpage pairs. The prompt were tailored to encourage questions that focus on visual concepts and can be answered using observable or associated information in the webpage content. Details of the prompts are shown in Table~\ref{tab:train-data-gen-prompt}.

\begin{table}[t]
\centering
\caption{Prompt used for FVQA-train VQA Generation.}
\label{tab:train-data-gen-prompt}
\begin{tabular}{p{1cm}|p{12cm}}
\toprule
\textbf{Usage} & \textbf{Prompt} \\ \midrule
Factual QA Generation & Your task is to generate a factual question–answer pair based on the given visual concept, the image, and the associated webpage content.

The generated question must strictly follow the requirements below:

1. The answer must contain the keyword visual concept.

2. The question must start with "Who", "What", or "Where".

3. The question must NOT include the visual concept itself, nor any background knowledge directly related to it.

4. The question should resemble something a curious human without prior knowledge about the image might ask.

In addition to the question, generate a concise and factual answer grounded in the visual concept, image, and webpage content.

Visual Concept: \texttt{\{visual\_concept\}}

Image: \texttt{\{image\}}

Webpage Content: \texttt{\{webpage\_content\}}

Respond only with the generated question and answer.
\\ \bottomrule
\end{tabular}
\end{table}

\subsection{Prompts for RL Training}
\label{appendix:train-prompts}
During training, prompts are used to guide and constrain the model’s behavior, ensuring that it interacts with the search tools in a consistent and structured manner. Specifically, we design three types of prompts for different points in the dialogue: one used at the beginning of the conversation, one after the model performs an image search, and another after it performs a text search. These context-specific prompts play a key role in aligning the model’s actions with the intended workflow and ensuring proper use of the multimodal search tools. Details of the prompts are shown in Table~\ref{tab:training-prompt}.
\begin{table}[H]
\centering
\caption{Prompts used during training at conversation start and after different search actions.}
\label{tab:training-prompt}
\begin{tabular}{p{1cm}|p{12cm}}
\toprule
\textbf{Usage} & \textbf{Prompt} \\ \midrule
1st Round & Answer the user's question based on the provided image. Examine the image carefully and identify any recognizable entities, such as faces, objects, locations, events, logos, or text. Determine whether you have sufficient knowledge to confidently recognize the main visual element and answer the user's question. If so, first explain your reasoning, then provide a clear and direct answer. 

If you are unable to confidently identify the visual element, stop and invoke the image search tool by appending the string \search{<search><img></search>} at the end of your response. This will trigger a Google Lens search using the original image to retrieve relevant information that can help you confirm the visual content.

Once you have sufficient visual understanding, combine it with the user's question and assess whether you can confidently answer. If so, answer the question directly using your own knowledge. If not, invoke the text search tool by generating a concise and specific query, and output it in the format \search{<text\_search>your query here</text\_search>} at the end of your response. Carefully craft your query to accurately retrieve the information needed to help answer the question. The text search tool will then use Google Search to return relevant information based on your query.

You must include your reasoning inside \reason{<reason>...</reason>} before taking any action, whether it is calling the image search tool, generating a text search query, or providing a final answer. The reasoning may involve analysis of the original image and question, interpretation of search results, or logical steps leading to the final answer.

All search results will be placed inside \information{<information>} and \information{</information>} and returned to you. When you are ready to answer the question, wrap your final answer between \answer{<answer>} and \answer{</answer>}, without detailed illustrations. For example: <answer>Titanic</answer>.

Here is the image and the question: \texttt{\{image\}\{question\}}  \\ \midrule
After image search & Original question: \texttt{\{question\}}

Please extract the visual element relevant to the user's question (such as faces, objects, locations, events, logos, or text) from the search results. Then, combine this information with the user's question and perform reasoning to determine whether a Google text search is needed to retrieve additional information to answer the question.

If a text search is needed, output the string \search{<text\_search>your query here</text\_search>} at the end of your response. Please generate a well-crafted query based on the visual element that will help retrieve the most relevant information.

If a text search is not needed, use your own knowledge to directly answer the user's question.

You must conduct your reasoning inside \reason{<reason>} and \reason{</reason>} before taking any action, whether it's generating a text search query or providing a final answer. If you decide to give the final answer, place it inside \answer{<answer>} and \answer{</answer>}, without detailed explanation or illustration. For example: <answer>Titanic</answer> \\ \midrule
After text search & Original question: \texttt{\{question\}}

Please analyze the search results and the user's question and continue reasoning inside \reason{<reason>} and \reason{</reason>}.

If you determine that additional knowledge is still required to answer the user's question, stop responding to the question and instead report a warning by outputting the string "Unable to answer due to lack of relevant information" at the end of your response.

If no further external information is needed, you should provide the final answer by placing it within \answer{<answer>} and \answer{</answer>}. The answer must be concise, clear, and to the point, without any additional explanation or elaboration. \\ 
\bottomrule
\end{tabular}
\end{table}

\subsection{Prompts for Webpage Summarization Model}
\label{appendix:web-summary-prompts}
The prompt shown in Table~\ref{tab:webpage-summary-prompt} is used to guide the webpage summarization model in the text search pipeline to effectively extract and summarize webpage content based on the given user question. The goal is to ensure that the returned content is more focused and concise, thereby reducing overall token consumption.

\begin{table}[t]
\centering
\caption{Prompt used for Webpage Summarization Model.}
\label{tab:webpage-summary-prompt}
\begin{tabular}{p{1cm}|p{12cm}}
\toprule
\textbf{Usage} & \textbf{Prompt} \\ \midrule
System Message & You are a helpful assistant. Your task is to summarize the main content of the given web page in no more than five sentences. Your summary should cover the overall key points of the page, not just parts related to the user's question. \\ \midrule
Prompt & If any part of the content is helpful for answering the user's question, be sure to include it clearly in the summary. Do not ignore relevant information, but also make sure the general structure and main ideas of the page are preserved. Your summary should be concise, factual, and informative. 

Webpage Content (first 30000 characters) is: \texttt{\{webpage\_content\}}

Question: \texttt{\{question\}}
\\ \bottomrule
\end{tabular}
\end{table}

\subsection{Prompt for Direct Answer Baseline}
\label{appendix:direct-answer-prompts}
Table~\ref{tab:direct-answer-prompt} presents the prompt template used in the Direct Answer baseline. In this setting, the model is expected to provide a concise answer to a given question based solely on the visual content of the image.

\begin{table}[t]
\centering
\caption{Prompt used for Direct Answer baselines.}
\label{tab:direct-answer-prompt}
\begin{tabular}{p{1cm}|p{12cm}}
\toprule
\textbf{Usage} & \textbf{Prompt} \\ \midrule
1st-Round & Based on the image, answer the question with as few words as you can. 

Question: \texttt{\{question\}} Image: \texttt{image}
\\ \bottomrule
\end{tabular}
\end{table}

\subsection{Prompts for RAG Workflow Baseline}
\label{appendix:rag-prompts}
Under the RAG workflow, models are required to perform exactly two search operations for each VQA example using our multimodal search tools: first an image search, followed by a text search. Under the RAG workflow, models are required to perform exactly two search operations for each VQA example using our multimodal search tools: first an image search, followed by a text search. This results in a maximum of three conversation rounds. Specifically, given an input image and question, the model is presented with the image search results and the original question in the first round and is prompted to generate a relevant text query to aid in answering. In the second round, the retrieved text results based on that query are provided to the model, which is then asked to generate the final answer. In our experimental setup, the RAG workflow reaches the maximum allowed number of search steps, making it a strong baseline that provides the model with the most external information. Details of the prompts of RAG Workflow are shown in Table~\ref{tab:rag-prompt}.
\begin{table}[t]
\centering
\caption{Prompts used for RAG workflow baselines. Text in \search{red} indicates content returned by the multimodal search tool.}
\label{tab:rag-prompt}
\begin{tabular}{p{1cm}|p{12cm}}
\toprule
\textbf{Usage} & \textbf{Prompt} \\ \midrule
1st Round & You are given a question accompanied by an image that cannot be answered without external knowledge. To assist your understanding, you are also provided with image search results consisting of five web pages related to the original image, ranked by relevance. Each result includes the main image from the web page and its title.

Question: \texttt{\{question\}} Image: \texttt{\{image\}}

\search{Image Search Results: }

\search{1. Webpage Image: \texttt{\{image\}} Webpage Title: \texttt{\{title\}}}

\search{2. Webpage Image: \texttt{\{image\}} Webpage Title: \texttt{\{title\}}}

\search{...}

\search{5. Webpage Image: \texttt{\{image\}} Webpage Title: \texttt{\{title\}}}

Assume you have access to a search engine (e.g., google). Based on the question, image and image search results, please raise a text query to the search engine to search for what is useful for you to answer the question correctly. You need to consider the characteristics of asking questions to search engines when formulating your questions. Now give the text query to search engine directly (do not wrap it in quotes or add any explanation): \\ \midrule
2nd Round & You should now read the text search result and answer the answer the question based on the image provided. 

\search{Text Search Results:}

\search{...}

Original question: \texttt{\{question\}}

Answer the question with as few words as you can.
\\ \bottomrule
\end{tabular}
\end{table}

\subsection{Prompts for LLM-as-Judge Evaluation}
\label{appendix:eval-prompts}
In our experiments, we adopt the LLM-as-judge approach to evaluate whether a model's response aligns with the ground truth answer. This method provides a more objective and generalizable way to assess performance on open-ended VQA tasks. Table~\ref{tab:eval-prompt} shows the full prompt used with GPT-4o, which serves as the judge LLM throughout all our experiments. Once GPT-4o determines whether each response is correct or incorrect, its judgments are used to compute the accuracy metric.
\begin{table}[t]
\centering
\caption{Full prompt used for GPT-4o as the judge LLM in all experiments.}
\label{tab:eval-prompt}
\begin{tabular}{p{1cm}|p{12cm}}
\toprule
\textbf{Usage} & \textbf{Prompt} \\ \midrule
System Message & You are an AI assistant tasked with evaluating the correctness of model responses based on an image, question, and ground truth answer. Your judgment should follow these principles:

1. Consider the image, question, and ground truth answer holistically before evaluating the model's response.

2. Your decision should be strictly \textbf{Yes or No}, based on whether the model's response is factually accurate and aligns with the ground truth answer.

3. If the model response is a more specific form of the ground truth answer, it is correct.

4. If the model response includes all key information but adds minor details, it is correct as long as the extra details are factually correct.

5. If the model response contradicts, modifies, or omits critical parts of the answer, it is incorrect.

6. For numerical values, ensure correctness even when presented in different units.

7. For names, check for first and last name correctness. If the middle name is extra but correct, consider it correct.

8. For yes/no questions, the response must exactly match "Yes" or "No" to be correct.

9. If the judgment can be made based solely on the text, you may choose to ignore the input image, as some images may be unfamiliar to you and could affect your judgment. Refer to the image only when necessary to minimize misjudgment.

10. If there are multiple candidate answers, you can also evaluate the model's response against all of them. If the response aligns with at least one candidate according to the rules above, it should be considered correct. 

Your output must be in the following format:

<judge>Yes/No</judge>

<reason>Explanation of why the answer is correct or incorrect.</reason> \\ \midrule
Prompt & \textbf{Image, Question, and Model Response Evaluation}

Question: \texttt{\{question\}}

Ground Truth Answer: \texttt{\{ground truth answer\}}

Candidate Answers: \texttt{\{candidate answers\}}

Model Response: \texttt{\{model response\}}

\textbf{Evaluation Instructions}

Evaluate whether the Model Response is correct based on the Image, Question, Ground Truth Answer and Candidate Answers. Follow the predefined judgment rules and provide a clear \textbf{Yes/No} answer along with a justification.

\textbf{Output Format}

<judge>Yes/No</judge>

<reason>Detailed reasoning following the evaluation principles.</reason>
\\ \bottomrule
\end{tabular}
\end{table}

\section{Details of Benchmark Datasets}
\label{appendix:benchmark-details}
\textbf{FVQA-test} As described in Section~\ref{method:data}, the \texttt{FVQA-test} set is a manually curated test set consisting of 1800 examples collected from three sources:
(1) 600 samples drawn from \texttt{FVQA-auto-vc}, with no overlap with the training set; each example was manually verified for correctness;
(2) 600 samples selected from the InfoSeek Human Split, where we manually re-annotated the answers, as the original human-labeled answers were not publicly released;
(3) 600 samples newly collected by human annotators.
This dataset covers a wide range of visual and textual knowledge categories, enabling a comprehensive evaluation of model performance.

\textbf{InfoSeek} InfoSeek~\cite{dataset/infoseek} is constructed through a semi-automated process that transforms Wikidata triples into natural language questions using human-authored templates. Annotators design question templates for 300 Wikidata relations, incorporating placeholders for visual entity types and units to ensure clarity. These questions are then paired with corresponding images and answers to form {image, question, answer} triplets. To ensure diversity and answerability, question-answer pairs that lack supporting evidence in Wikipedia are filtered out, and balanced sampling is applied across entities and relations. This design makes InfoSeek particularly suitable for evaluating information-seeking capabilities, as it emphasizes diverse, fact-based queries grounded in real-world knowledge and multimodal contexts. We randomly sampled 2000 examples from its test split due to the large dataset size.

\textbf{MMSearch} MMSearch~\cite{vision-rag/mmsearch} contains 300 manually collected examples spanning 14 subdomains, categorized into two types: News, which includes up-to-date events from August 2024 to ensure no overlap with LMM training data, and Knowledge, which focuses on rare, verified knowledge that current state-of-the-art LMMs (e.g., GPT-4o, Claude 3.5) fail to answer. Among these, 171 examples are visual questions (i.e., questions paired with images), while the remaining are purely textual. We use the 171 visual questions as our evaluation subset. This dataset is particularly well-suited for evaluating models' real-world information-seeking abilities, especially in scenarios that require up-to-date retrieval or reasoning over rare knowledge.

\textbf{SimpleVQA} SimpleVQA~\cite{dataset/simplevqa} is built from two main sources of seed examples. First, factual image-question pairs are selected from existing VQA datasets that meet real-world knowledge standards. These datasets were constructed after 2023 and focus on practical, application-driven content. Secondly, additional data is sourced via internet search, with expert annotators generating QA pairs based on retrieved images and factual content. These examples span a wide range of entities and events, with answers focused on objective, fact-based information involving entity recognition and attribute extraction. Building on these seed samples, the dataset further undergoes reliable difficulty filtering and human-in-the-loop quality control, making it a strong candidate for evaluation of models' factual and knowledge-based reasoning abilities. The final SimpleVQA benchmark contains 2,025 examples. To avoid language-related confounding factors, we select the 1,013 English QA pairs for evaluating model performance.

\textbf{LiveVQA} LiveVQA~\cite{dataset/livevqa} is built from six globally recognized news platforms, such as CNN, BBC, Yahoo, Forbes, AP News, and Variety, to ensure diversity and timeliness across visual and textual content. The dataset covers 14 major news categories, including sports, entertainment, science, economy, and health, and contains 3,602 VQA pairs. For each article, question–answer pairs are generated using GPT-4o, including both basic visual questions and more complex multi-hop questions that require reasoning over the accompanying text. This setup supports evaluation of real-world information-seeking and multimodal reasoning capabilities. 

\section{Details of Experimental Implementation}
\label{appendix:exp-setups}

\subsection{RL Training Setting}
\label{appendix:exp-training-hyperparameters-rl}
We adopt the Qwen2.5-VL-7B model as the backbone model, and the overall GRPO training is implemented based on the veRL~\cite{code/verl} framework. The training set is the \texttt{FVQA-train} dataset, which consists of approximately 1,600 search-free samples and 3,400 search-required samples. Training is conducted on 4 nodes with 8 Nvidia H100 GPUs each. The total batch size is set to 512, with a mini-batch size of 128. For each training prompt, we generate 8 rollouts, and each rollout allows up to 3 tool calls. Regarding the reward components, the search penalty and the weights for the format score are both set to 0.1. The learning rate is 2e-6, and we fix the KL divergence coefficient $\beta$ to 0.001 and the clip ratio $\epsilon$ to 0.2. For MMSearch-R1-7B, we use the checkpoint from step 50 (when training converges) for downstream evaluation.
\subsection{SFT Training Setting}
\label{appendix:exp-training-hyperparameters-sft}
For the SFT model used in the RL vs. SFT performance comparison experiments presented earlier. We fine-tune the Qwen2.5-VL-7B model using LLaMA-Factory~\cite{code/llama-factory}, with a dataset of 8,000 samples generated by prompting GPT-4o to perform on-demand search. Each sample contains up to three rounds of dialogue. The GPT-4o responses are used as supervision signals, while the user question in the first turn and tool-returned content in later turns are masked out during loss computation.
Fine-tuning is performed on a single machine with 8 Nvidia H100 GPUs. We use a per-device batch size of 1 and apply gradient accumulation over 2 steps. The model is trained for 1 epochs with a learning rate of 1e-5. A cosine learning rate scheduler is adopted, with 10\% of the total training steps used for warm-up.
\subsection{Evaluation Setting}
\label{appendix:exp-eval-setup}
We conduct model inference using the veRL framework by setting the $val\_only$ parameter of the trainer to True. The inference engine is based on vLLM, with $top\_p$ set to 1.0 and temperature set to 0. A single response is generated for each sample in the test set.

For evaluation, we adopt an LLM-as-Judge approach. Specifically, we use GPT-4o-20241120 as the judge model and apply a predefined prompt template (as shown in Table~\ref{tab:eval-prompt}) to assess each inference result. During judgment, $top\_p$ is set to 0.1 and temperature is set to 0. The model’s binary output ("Yes" or "No") is used to compute the final accuracy metric.

\section{Full Experimental Results}
\label{appendix:full-exp}
\subsection{Evaluation on General VQA Benchmarks}
To assess whether RL training affects general VQA capabilities, we compare MMSearch-R1-7B with its backbone model Qwen2.5-VL-7B on a suite of standard VQA benchmarks, including AI2D~\cite{bench/ai2d}, ChartQA~\cite{bench/chartqa}, LLaVA-Wilder~\cite{bench/llava-wilder}, MathVista~\cite{bench/mathvista}, MME~\cite{bench/mme}, and OCRBench~\cite{bench/ocrbench}. All experiments are conducted using LMMS-Eval~\cite{code/lmms-eval}. As shown in Table~\ref{tab:general-vqa}, MMSearch-R1-7B achieves comparable performance across all benchmarks, slightly outperforming the base model on LLaVA-Wilder and MathVista while maintaining similar results on AI2D, ChartQA, OCRBench and MME. These results suggest that our reinforcement learning process, while enhancing search-related behavior, preserves the model’s general visual understanding and reasoning ability.
\begin{table}[t]
\centering
\caption{Performance Comparison on General VQA Benchmarks}
\label{tab:general-vqa}
\setlength{\tabcolsep}{4pt}
\begin{tabular}{c|cccccc}
\toprule
\multirow{2}{*}{Model} & AI2D & ChartQA & LLaVA-Wilder & MathVista & MME & OCRBench \\
 & test & test & small & testmini & test & - \\ \midrule
Qwen2.5-VL-7B & 93.0 & 86.6 & 73.5 & 68.2 & 623/1705 & 84.7 \\
MMSearch-R1-7B & 92.8 & 86.1 & 74.8 & 68.8 & 622/1707 & 84.5 \\ \bottomrule
\end{tabular}
\end{table}

\subsection{Ablation Study on Reward Modeling}
\label{appendix:abl-4o-reward}

\begin{table}[t]
\footnotesize
\centering
\caption{Ablation Study on Reward Modeling. 
"Acc (\%)" denotes the accuracy evaluated by LLM-as-Judge, while "SR (\%)" represents the search ratio, defined as the percentage of total search calls made relative to the maximum allowed search steps for each method. \textbf{$^*$EM} indicates that the accuracy score used in the reward function is computed via exact string match (EM), while \textbf{$^*$4o} denotes that correctness is judged by GPT-4o.}
\label{tab:4o-reward}
\setlength{\tabcolsep}{5pt}
\begin{tabular}{ccccccccccccc}
\toprule
\multicolumn{1}{c|}{\multirow{3}{*}{\textbf{Model}}} & \multicolumn{2}{c|}{\multirow{2}{*}{\textbf{Average}}} & \multicolumn{4}{c|}{\textbf{In-Domain}} & \multicolumn{6}{c}{\textbf{Out-of-Domain}} \\
\multicolumn{1}{c|}{} & \multicolumn{2}{c|}{} & \multicolumn{2}{c}{\textbf{FVQA-test}} & \multicolumn{2}{c|}{\textbf{InfoSeek}} & \multicolumn{2}{c}{\textbf{MMSearch}} & \multicolumn{2}{c}{\textbf{SimpleVQA}} & \multicolumn{2}{l}{\textbf{LiveVQA}} \\
\multicolumn{1}{c|}{} & \textbf{Acc} & \multicolumn{1}{c|}{\textbf{SR}} & \textbf{Acc} & \textbf{SR} & \textbf{Acc} & \multicolumn{1}{c|}{\textbf{SR}} & \textbf{Acc} & \textbf{SR} & \textbf{Acc} & \textbf{SR} & \multicolumn{1}{c}{\textbf{Acc}} & \multicolumn{1}{c}{\textbf{SR}} \\ \midrule
\multicolumn{13}{c}{\textit{\textbf{RAG Workflow}}} \\ \midrule
\multicolumn{1}{c|}{GPT4o} & 62.1 & \multicolumn{1}{c|}{100} & 66.0 & 100 & 59.1 & \multicolumn{1}{c|}{100} & 62.5 & 100 & 63.4 & 100 & 59.6 & 100 \\
\multicolumn{1}{c|}{Gemini 2.5 Pro} & 61.8 & \multicolumn{1}{c|}{100} & 66.1 & 100 & 56.7 & \multicolumn{1}{c|}{100} & 62.5 & 100 & 65.9 & 100 & 57.8 & 100 \\
\multicolumn{1}{c|}{Qwen2.5-VL-72B} & 59.6 & \multicolumn{1}{c|}{100} & 62.2 & 100 & 59.4 & \multicolumn{1}{c|}{100} & 59.6 & 100 & 61.0 & 100 & 56.0 & 100 \\
\multicolumn{1}{c|}{Qwen2.5-VL-32B} & 55.1 & \multicolumn{1}{c|}{100} & 57.0 & 100 & 56.8 & \multicolumn{1}{c|}{100} & 57.9 & 100 & 54.5 & 100 & 49.6 & 100 \\
\multicolumn{1}{c|}{Qwen2.5-VL-7B} & 51.6 & \multicolumn{1}{c|}{100} & 52.9 & 100 & 53.7 & \multicolumn{1}{c|}{100} & 52.2 & 100 & 51.6 & 100 & 48.0 & 100 \\ \midrule
\multicolumn{13}{c}{\textit{\textbf{On-demand Search}}} \\ \midrule
\multicolumn{1}{c|}{\makecell{MMSearch-R1-7B\\\textbf{$^*$EM}}} & 55.7 & \multicolumn{1}{c|}{77.3} & 59.1 & 83.4 & 57.1 & \multicolumn{1}{c|}{68.3} & 55.6 & 91.8 & 58.1 & 56.8 & 48.5 & 86.2 \\
\multicolumn{1}{c|}{\makecell{MMSearch-R1-7B\\\textbf{$^*$4o}}} & 59.5 & \multicolumn{1}{c|}{82.6} & 62.1 & 88.9 & 59.1 & \multicolumn{1}{c|}{81.8} & 61.9 & 92.7 & 59.8 & 72.6 & 54.7 & 76.9 \\ \bottomrule
\end{tabular}
\end{table}

Our main training framework adopts a rule-based reward design based on exact string match, which is simple, deterministic, and scalable. However, this formulation may not fully capture the semantic correctness of answers, especially when multiple surface forms express the same factual content. To explore whether a more flexible and context-aware reward signal can better guide model learning, we conduct an additional experiment using GPT-4o as the reward model during training. GPT-4o provides semantic-level accuracy judgments, allowing us to assess the impact of richer reward feedback on model behavior and final performance. The prompt used to elicit reward signals from GPT-4o is the same as the one used in our evaluation setup, as shown in Table~\ref{tab:eval-prompt}.

As shown in Table~\ref{tab:4o-reward}, training with GPT-4o reward leads to a clear improvement across all datasets. The \textbf{$^*$4o} version achieves an average accuracy of 59.5\%, which is 3.8 points higher than the \textbf{$^*$EM} version. These results indicate that more general reward supervision holds greater potential for open-ended tasks. GPT-4o helps avoid false negatives caused by exact match and leads to improved robustness. However, it may introduce bias, higher training cost and struggles with questions beyond its knowledge scope. We believe that exploring more general reward modeling represents a promising direction for future research. Notably, this experiment is conducted on a subset of the \texttt{FVQA-train} dataset, in which search-free examples are relatively underrepresented. As a result, the model exhibits a higher average search rate at convergence compared to Table~\ref{tab:finding-1}.

\section{Limitations}
\label{appendix:limitations}
While MMSearch-R1 demonstrates strong performance and introduces a practical framework for on-demand multimodal search, several limitations remain.

First, the interaction between the model and external multimodal search tools still has room for improvement in terms of stability and quality. For example, image search currently requires submitting the full image, which may not be optimal for retrieving localized visual content. The text search pipeline is composed of several components, including SerpAPI, Jina Reader, and a summarization model. Each component may introduce potential sources of variability. Specifically, the ranking of result returned by SerpAPI may vary over time; Jina Reader may fail to extract content from certain domains due to restrictions or errors; and the summarizer may generate hallucinations when extracting relevant information. Despite iterative refinement of the pipeline, our monitoring during training revealed an end-to-end failure rate of approximately 0.2\% for image search and 1\% for text search, where failure is defined as receiving no valid results. These numbers are higher when considering partial failures, such as retrieving fewer than the expected top five results. As training scales up, ensuring stability and consistency in tool outputs remains a nontrivial challenge.

Second, our reward design based on exact string match, while simple and scalable, has limited flexibility. This design is well-suited for short factual questions with unambiguous answers. However, it may penalize answers that are semantically correct but differ slightly in phrasing. This limitation affects the generalization of the reward function to more complex or open-ended QA tasks. Although our dataset focuses on fact-based QA that aligns well with this evaluation strategy, more flexible reward signals could help expand the framework to broader question types and reduce reliance on surface-form matching. To explore this, we conducted a preliminary experiment comparing EM-based rewards with GPT-4o-based rewards, which showed that the latter offers better tolerance to answer variation and may support more nuanced evaluation (see Appendix~\ref{appendix:abl-4o-reward} for details). 

In general, improving the robustness of tool interactions and enhancing the expressiveness of the reward function will be essential for scaling MMSearch-R1 to more adaptive and reliable multimodal reasoning agents.

\section{Broader Impacts}
\label{appendix:broader-impacts}
Our work focuses on improving the ability of multimodal models to access and reason over external information through search. Although this capability has clear benefits for building more informative and adaptive assistants, it also introduces potential risks. For example, models that autonomously retrieve and summarize web content may surface outdated, biased, or misleading information. Additionally, over-reliance on real-time search may raise concerns around content verifiability, copyright, or inadvertent propagation of misinformation.

To mitigate these potential risks, we recommend careful filtering of the retrieved content, attribution of information sources, and incorporating mechanisms that allow users to trace and verify the provenance of model-generated responses. We also encourage future research on safer retrieval strategies, such as domain whitelisting and adaptive uncertainty calibration, especially in high-stakes applications.

\section{Examples of FVQA Dataset}
\label{appendix:example-fvqa}
\begin{figure}[p]
  \centering
  \includegraphics[width=0.96\textwidth]{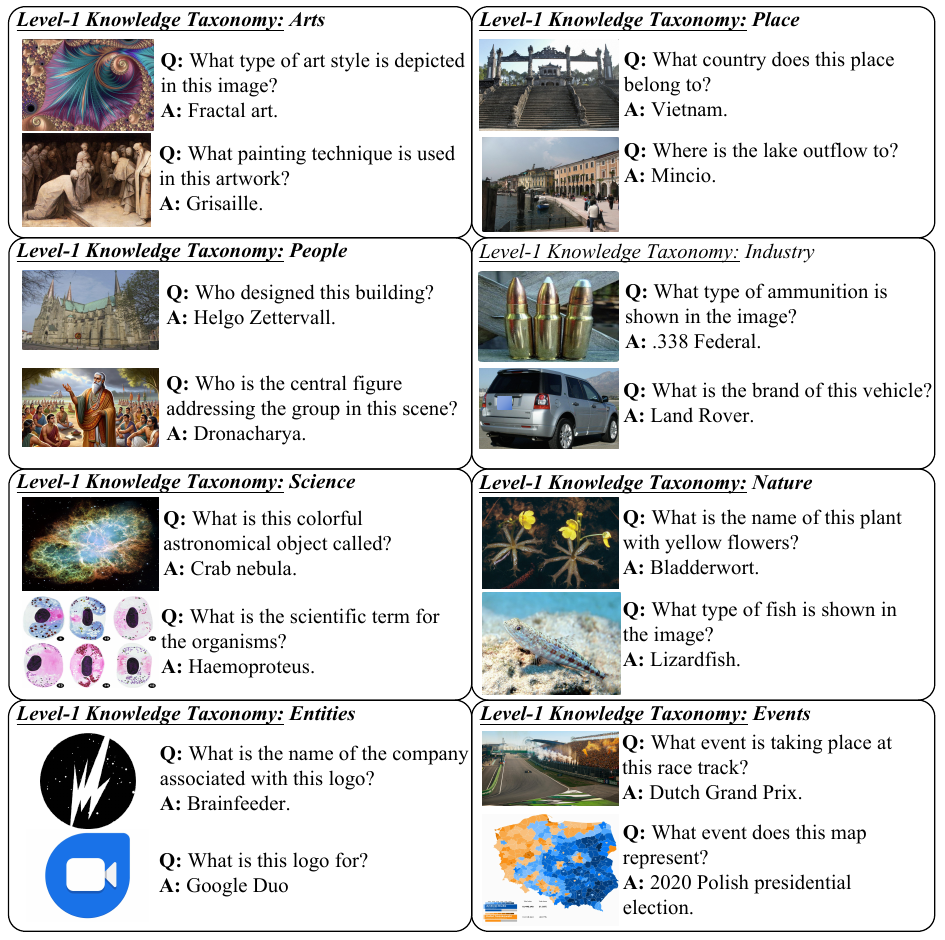}
  \caption{Examples of \texttt{FVQA} dataset.}
  \label{fig:example_fvqa}
\end{figure}

\begin{figure}[p]
  \centering
  \includegraphics[width=0.96\textwidth]{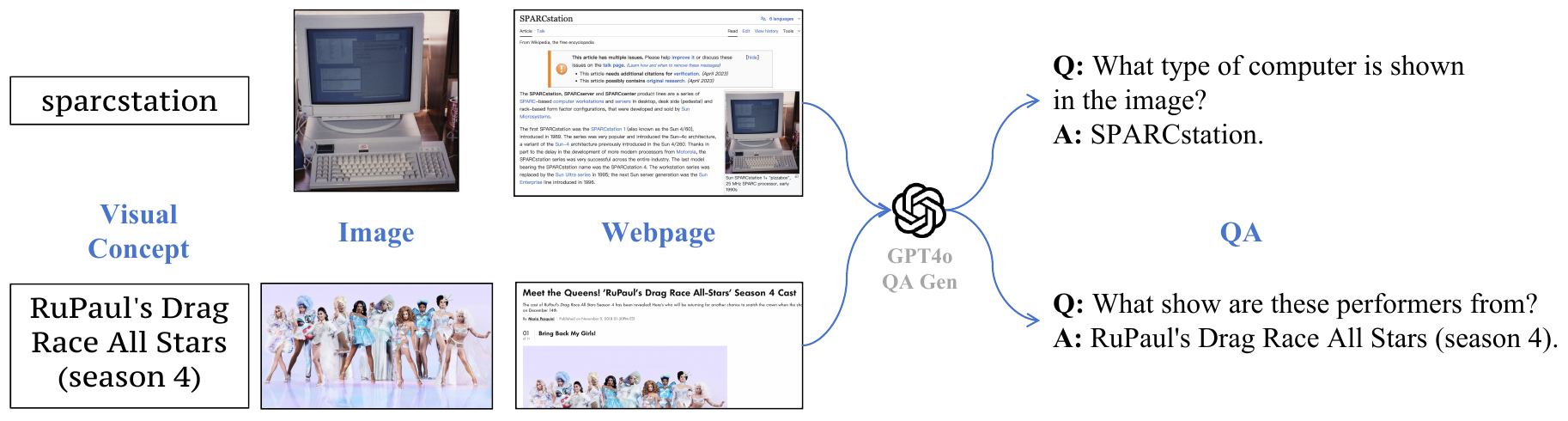}
  \caption{Examples of \texttt{FVQA-auto-vc} QA Generation.}
  \label{fig:example_fvqa_auto_vc_gen}
\end{figure}

To better illustrate the characteristics of our dataset, we present representative examples from the \texttt{FVQA} dataset in Figure~\ref{fig:example_fvqa}.  QA pairs are designed to be knowledge-intensive and fact-oriented, covering a broad range of visual and textual knowledge types. We also present examples from the data construction process of \texttt{FVQA-auto-vc} in Figure~\ref{fig:example_fvqa_auto_vc_gen}.

\section{Case Studies}
\begin{figure}[p]
  \centering
  \includegraphics[width=0.99\textwidth]{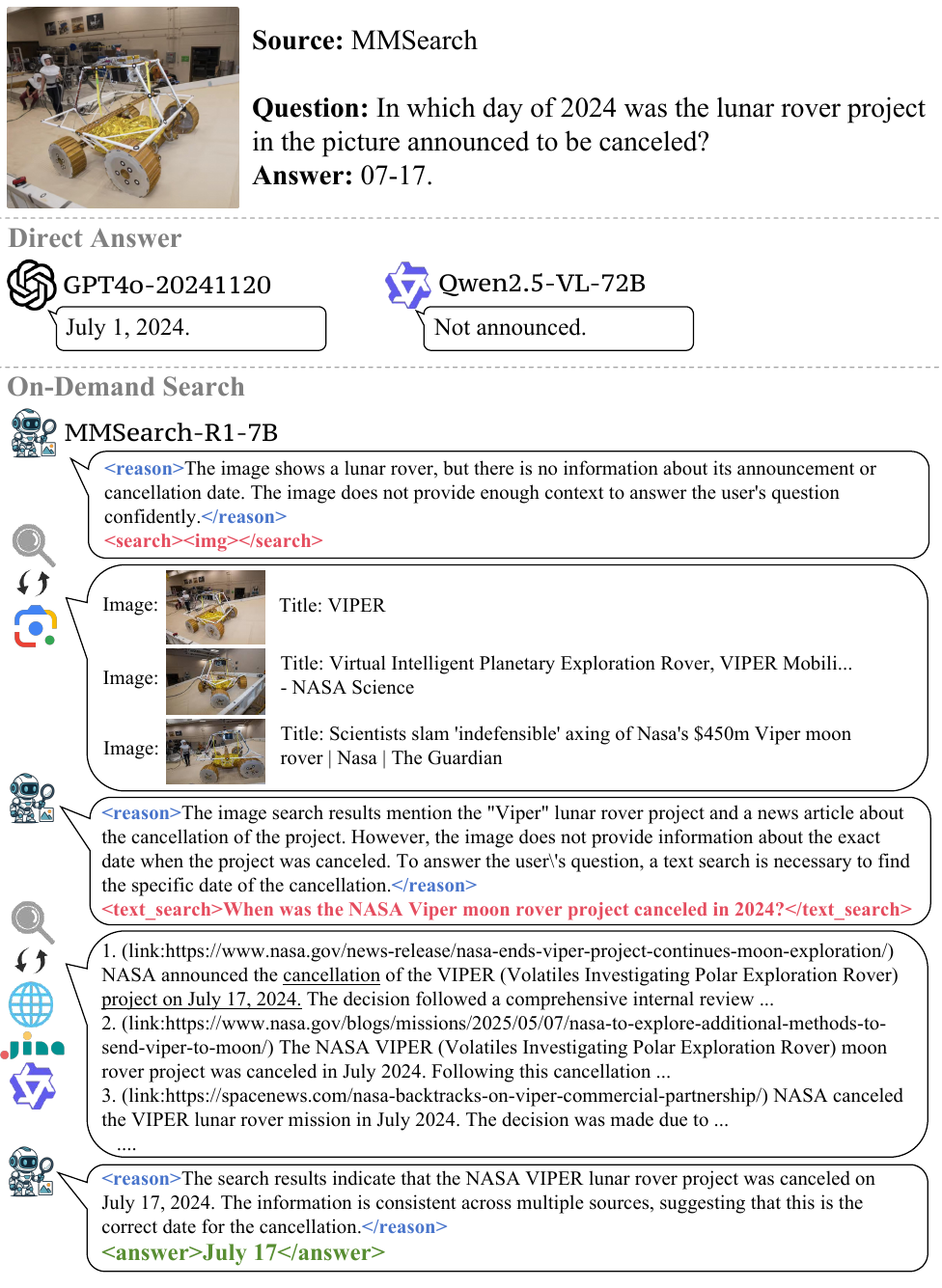}
  \caption{Case study 1 of MMSearch-R1.}
  \label{fig:case_study_1_msmearch_viper}
\end{figure}

\begin{figure}[p]
  \centering
  \includegraphics[width=0.99\textwidth]{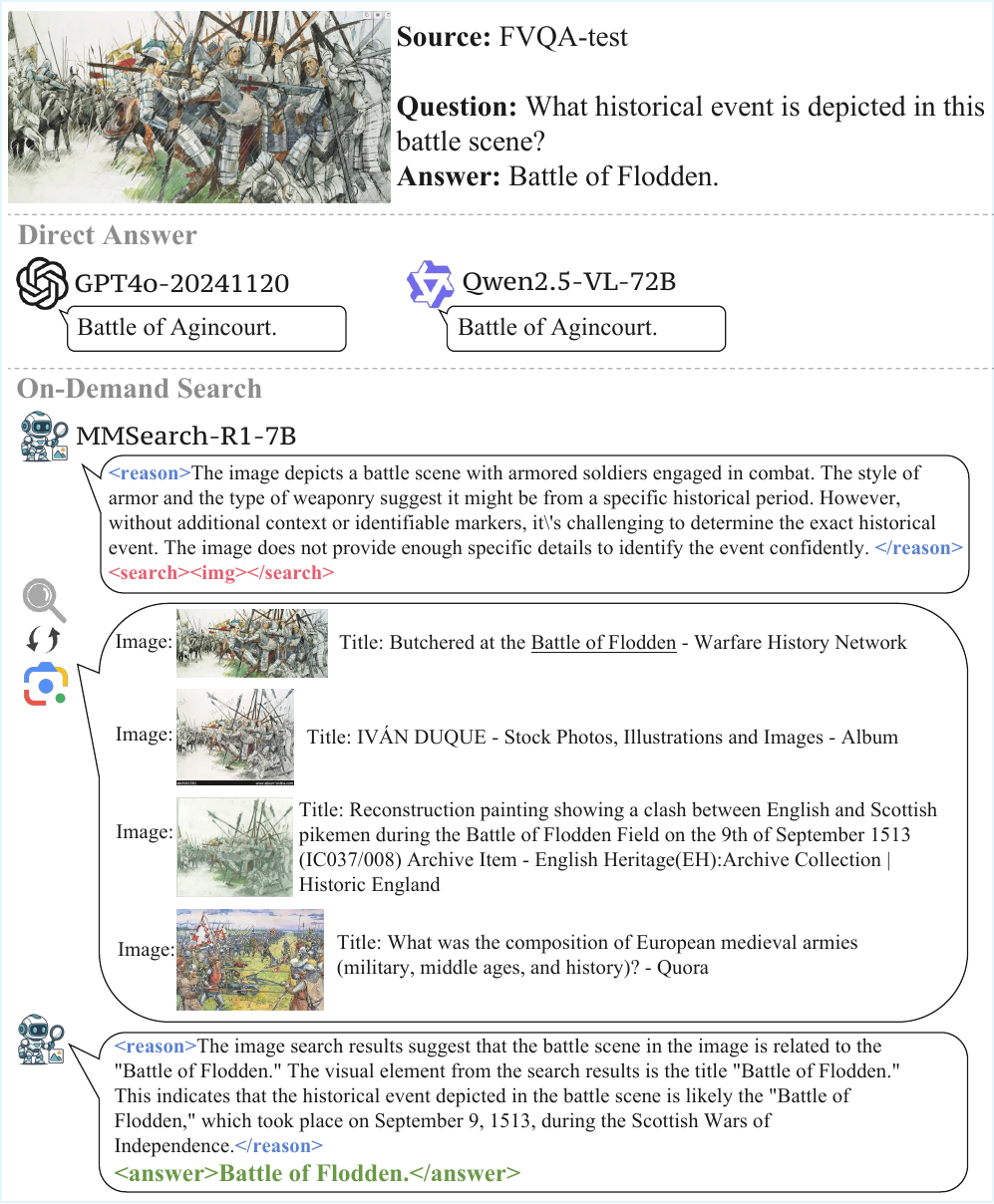}
  \caption{Case study 2 of MMSearch-R1.}
  \label{fig:case_study_3_fvqa_flodden}
\end{figure}

As shown in Figure~\ref{fig:case_study_1_msmearch_viper} and Figure~\ref{fig:case_study_3_fvqa_flodden}, we present several representative case studies to illustrate how MMSearch-R1 performs on complex, real-world information-seeking VQA tasks. These examples highlight the model’s ability to reason about when and how to invoke search tools, generate effective queries, and synthesize the retrieved information to arrive at accurate answers. For ease of formatting and readability, we omit some of the retrieved search results in certain cases.

\clearpage

\clearpage

\end{document}